\documentclass{article}

\PassOptionsToPackage{numbers}{natbib}

 \usepackage[main, final]{neurips_2026}

\usepackage[utf8]{inputenc} 
\usepackage[T1]{fontenc}    
\usepackage{hyperref}       
\usepackage{url}            
\usepackage{booktabs}       
\usepackage{amsfonts}       
\usepackage{nicefrac}       
\usepackage{microtype}      
\usepackage{xcolor}         
\usepackage{microtype}
\usepackage{graphicx}
\usepackage{subcaption}
\usepackage{booktabs}

\usepackage{amsmath}
\usepackage{amssymb}
\usepackage{mathtools}
\usepackage{amsthm}

\usepackage{algorithm}
\usepackage{algorithmic}

\usepackage[capitalize,noabbrev]{cleveref}
\usepackage{multirow}

\usepackage[textsize=tiny]{todonotes}
\usepackage{array}
\usepackage{enumitem}

\title{HandXFM: Semantic-Structural Distillation for Hand Radiograph Foundation Models}

\author{
\resizebox{0.97\textwidth}{!}{%
  \textbf{Yuxi Long}$^{1}$,
  \textbf{Ganlin Feng}$^{1}$,
  \textbf{Lianghong Chen}$^{1}$,
  \textbf{Liam J. O'Neil}$^{2}$,
  \textbf{Carol A. Hitchon}$^{3}$,
  \textbf{Pingzhao Hu}$^{1,4,*}$%
}
\\[0.5em]
$^{1}$Department of Computer Science, Western University, London, ON, Canada
\\
$^{2}$Department of Medicine, University of Manitoba, Winnipeg, MB, Canada
\\
$^{3}$Department of Internal Medicine, University of Manitoba, Winnipeg, MB, Canada
\\
$^{4}$Department of Biochemistry, Western University, London, ON, Canada
\\[0.3em]
\texttt{ylong66@uwo.ca}, \texttt{phu49@uwo.ca}
\\[0.2em]
}

\begin{document}

\maketitle

\begingroup
\renewcommand{\thefootnote}{*}
\footnotetext{Corresponding author. Department of Computer Science,
Western University, 1400 Western Road, London,
Ontario N6G 2V4, Canada. E-mail: phu49@uwo.ca (P.H.)}
\endgroup

\begin{abstract}
Hand radiographs support diverse clinical tasks, including skeletal maturity assessment, rheumatoid arthritis scoring, abnormality screening, and anatomical localization, yet existing models are typically trained separately for each task and dataset. We present HandXFM, a domain-specific foundation model that leverages cross-modality knowledge for comprehensive hand X-ray understanding. HandXFM distills semantic priors from BiomedCLIP and structural cues from a chest X-ray Vision Transformer (ViT) through a unified alignment framework, integrating complementary medical knowledge in an interpretable way. After pretraining, HandXFM can be adapted to a range of musculoskeletal tasks, including bone age estimation, SvH score prediction, abnormality detection, joint localization, bone segmentation, and visual question answering. It consistently outperforms task-specific and single-expert baselines, achieving up to 0.11 improvement in Pearson correlation coefficient (PCC) for bone age prediction and a 12\% accuracy gain in abnormality classification. Moreover, HandXFM generalizes to unseen datasets and yields cross-attention maps linking clinical terms to anatomical regions. These results demonstrate that multi-expert distillation effectively unifies semantic and structural supervision within a single pretrained model, establishing HandXFM as an interpretable and generalizable foundation model for hand radiograph analysis.
\end{abstract}

\section{Introduction}

Hand radiography is a cornerstone modality in musculoskeletal imaging, used to assess growth and development, diagnose arthritis and trauma, quantify joint damage, and guide treatment. Clinical tasks including bone age estimation, SvH (Sharp/van der Heijde) scoring~\cite{svh_score}, joint detection, bone segmentation, and abnormality classification rely on hand X-rays as structural evidence. Despite their importance, computational models for hand radiographs are still largely developed separately for each task and dataset, limiting transfer across institutions, supervision levels, and clinical objectives.

Existing approaches are mostly task-specific and annotation-dependent~\cite{bonet, anhui}, making it difficult to build a unified representation backbone that captures shared anatomical and radiographic patterns. In contrast to the rapid progress of large-scale foundation models in natural and multimodal imagery~\cite{radford2021learning, He_2022_CVPR}, the hand X-ray domain still lacks a reusable pretrained model tailored to its structure-dominant nature. Prior medical foundation models provide useful semantic priors~\cite{zhang2024biomedclip, wang2022medclip}, but often rely on cross-modal supervision or single-expert training, leading to limited structural reasoning and weaker generalization across datasets. They also tend to overlook fine-grained anatomical cues such as bone boundaries, joint gaps, and cortical contours, which are critical for hand radiograph interpretation.

To address these limitations, we present \textbf{HandXFM}, a domain-specific foundation model for hand X-rays. HandXFM integrates a transformer-based student encoder with expert teacher models, i.e., domain-specialized pretrained models. It adopts a multi-expert masked latent alignment framework~\cite{mlr} combining alignment and reconstruction objectives, edge-aware masking for structural boundaries, and an Exponential Moving Average (EMA) teacher~\cite{ema} for stable supervision. These design choices are motivated by key characteristics of hand radiographs: subtle boundary-sensitive findings, heterogeneous downstream supervision, and cross-dataset transfer.

Once pretrained, HandXFM can be adapted to new datasets as a backbone for downstream tasks including regression, classification, detection, segmentation, and VQA. It can be fine-tuned with minimal supervision, making it suitable for multi-center medical pipelines. Extensive experiments show that HandXFM outperforms prior self-supervised and foundation baselines across tasks and generalizes to unseen datasets such as the Digital Hand Atlas. Visualizations indicate that its attention aligns with clinically meaningful regions, supporting anatomical grounding and interpretability.

\textbf{Our main contributions are summarized as follows:}
\vspace{-0.4em}
\begin{itemize}[leftmargin=*, itemsep=0pt, topsep=0pt, parsep=0pt, partopsep=0pt]
    \item We curate a large multi-source hand X-ray dataset and propose HandXFM, a domain-specific foundation model for hand radiographs, capable of learning unified radiographic representations.
    \item We introduce a multi-expert masked latent alignment framework that fuses semantic and structural priors from experts through alignment, reconstruction, edge-aware masking, and EMA stabilization.
    \item We validate HandXFM across diverse downstream tasks, including regression, classification, detection, segmentation, and VQA, demonstrating transferability and anatomical interpretability.

\end{itemize}
\vspace{-0.3em}

\section{Related Work}
\label{sec:related_work}

Recent advances in large-scale visual pretraining have produced powerful foundation models such as DINOv3~\cite{siméoni2025dinov3}, MAE~\cite{He_2022_CVPR}, CLIP~\cite{radford2021learning}, and EVA~\cite{EVA}.
They learn general-purpose representations from natural images through contrastive or masked objectives, achieving strong transferability.
However, assumptions on color textures, open-world semantics, and text-aligned supervision limit their applicability to grayscale, structure-dominant medical domains.

Medical foundation models such as MedCLIP~\cite{wang2022medclip}, BiomedCLIP~\cite{zhang2024biomedclip}, and M3AE~\cite{chen2022m3ae} introduce semantic priors and cross-modal transfer~\cite{sun2025_vlmreview,radclip}.
Yet many rely on textual supervision~\cite{bluethgen2025_chestvlm} or single-domain training~\cite{paschali2025_fm,wu2025_radfm}, with limited anatomical or structure-aware reasoning.
Recent musculoskeletal and multi-anatomy X-ray foundation models, such as SKELEX~\cite{kim2026generalizable} and XR-0~\cite{singla2025multi}, improve radiographic transfer but are not tailored to fine-grained hand-specific reasoning.

Hand X-ray studies remain task-specific, covering bone age prediction~\cite{Halabi2019RSNA,andleeb2025}, rheumatoid arthritis scoring~\cite{Stolpovsky_2023_ICCV,moradmand2025}, and abnormality classification~\cite{rajpurkar2018mura,alzubaidi2024_shoulder}.
These CNN- or ViT-based models perform well within individual domains but generalize poorly across datasets and tasks.
To the best of our knowledge, no foundation model has been developed specifically for hand X-rays, and no existing approach unifies multi-task hand radiography in a single framework.

Unlike prior methods, HandXFM bridges semantic foundation models and anatomical hand analysis.
It combines multi-expert latent alignment, edge-aware masking, and EMA consistency to learn unified, structure-informed representations from unlabeled hand X-rays.


\section{Method}
\label{sec:method}

\begin{figure}[t]
  \centering
  \includegraphics[width=\textwidth]{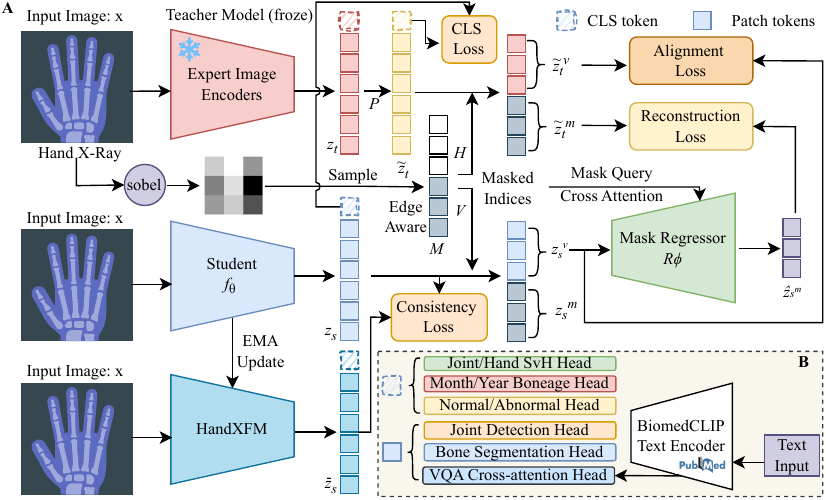}
  \caption{
  \textbf{Overview of the proposed HandXFM framework.}
  \textbf{(A) Pretraining architecture.}
  HandXFM jointly aligns the student encoder with multiple frozen medical experts while reconstructing masked latent features through the regressor.
  Edge-aware masking emphasizes structurally salient regions using Sobel gradients.
  The student learns from complementary objectives: multi-expert alignment loss, masked latent reconstruction loss, and EMA-based consistency loss, with an additional CLS-level supervision to preserve global semantics.
  \textbf{(B) Downstream adaptation.}
  After pretraining, the resulting foundation model serves as a universal encoder for diverse radiographic downstream tasks, including bone segmentation, joint detection, normal/abnormal classification, bone age prediction, SvH scoring, and VQA.
  The pretrained representation provides strong anatomical priors that transfer effectively across tasks and datasets.
  }
  \label{fig:overview}
  \vskip -0.2in
\end{figure}

HandXFM learns anatomically meaningful representations from unlabeled hand radiographs.
Instead of relying on a single self-supervised objective, it combines visual and semantic supervision within a unified pretraining framework.
The model aligns its latent space with multiple medical experts, reconstructs occluded features through contextual prediction, and maintains stable representation dynamics over time.
An overview of the HandXFM framework is illustrated in \Cref{fig:overview}. 

\subsection{Preliminary}

Given an unlabeled hand X-ray image $x \in \mathbb{R}^{3\times224\times224}$, preprocessed as described in the Supp.\ref{supp:preprocess}, the student encoder $f_{\theta}$ produces a set of latent embeddings:
\begin{equation}
z_s = f_{\theta}(x) = [z_{\mathrm{cls}}; z_1, z_2, \dots, z_N], \quad z_i \in \mathbb{R}^D ,
\end{equation}
where $z_{\mathrm{cls}}$ represents the global token and $\{z_i\}$ denote patch-level embeddings.
Following DINOv3 \cite{siméoni2025dinov3}, the image is divided into $N=14\times14=196$ patches. Frozen medical experts $\{t_j\}_{j=1}^{J}$, each pretrained on different imaging domains, provide complementary latent targets:
\begin{equation}
z_{t_j} = t_j(x) = [z^{(j)}_{\mathrm{cls}}; z^{(j)}_1, \dots, z^{(j)}_N].
\end{equation}

Each frozen expert feature is mapped into the student feature space by a learnable linear projection, anchoring supervision to pretrained medical representations while introducing complementary semantic and structural cues:
\begin{equation}
\tilde{z}^{(j)} = P_j z_{t_j}, \quad P_j : \mathbb{R}^{D_j} \rightarrow \mathbb{R}^{D}.
\end{equation}

To encourage latent prediction, a binary mask
$M\in\{0,1\}^{N}$ is sampled per image.
The visible subset
$\mathcal{V}=\{i\mid M_i=0\}$
is used for alignment, while the masked subset
$\mathcal{H}=\{i\mid M_i=1\}$
is used for reconstruction.

\subsection{Multi-Expert Alignment}
\label{sec:encoder_alignment}

The alignment objective enables the student to absorb knowledge from multiple medical experts and form a domain-agnostic representation space.
Each expert contributes a distinct prior: a text–vision model such as BiomedCLIP~\cite{zhang2024biomedclip} provides semantic abstraction, while a radiographic ViT~\cite{hf_vit_cxr} offers radiology-typical spatial–tonal cues.
By jointly aligning to these signals, the student can encode both conceptual and style consistency. Visible patch tokens align with the corresponding expert projections through a mean-squared latent alignment loss:
\begin{equation}
\mathcal{L}_{\mathrm{align}}
= \frac{1}{|\mathcal{V}|}
  \sum_{i \in \mathcal{V}}
  \frac{1}{J}\sum_{j=1}^{J}
  \|\, z_i - \tilde{z}^{(j)}_i \,\|_2^2 .
\label{eq:align}
\end{equation}
This encourages the model to preserve local relationships that remain semantically consistent across domains. To maintain global coherence, we distill the image-level [CLS] token from all experts:
\begin{equation}
\mathcal{L}_{\mathrm{cls}}
= \frac{1}{J}
  \sum_{j=1}^{J}
  \|\, z_{\mathrm{cls}} - \tilde{z}^{(j)}_{\mathrm{cls}} \,\|_2^2 .
\label{eq:cls}
\end{equation}
Together, these objectives transfer both fine-grained spatial correspondences and holistic semantics into the student encoder. By jointly attending to multiple experts, the model learns to represent hand anatomy agnostic to dataset origin or modality differences. The resulting latent space is both semantically organized and structurally smooth, allowing consistent mapping of similar joints, bones, and pathological cues in varying imaging conditions. This multi-expert alignment allows HandXFM to generalize to diverse downstream tasks without task-specific supervision.

\subsection{Masked Latent Reconstruction}
\label{sec:reconstruction}

To complement expert alignment, HandXFM performs token-level masked latent reconstruction by withholding a subset of encoder output tokens and predicting their expert-aligned targets from the remaining tokens. This objective regularizes the student toward spatial continuity and anatomical dependencies, rather than serving as input-level masked image modeling.

Let $z^v_s=\{z_i\mid i\in\mathcal{V}\}$ denote visible patch embeddings and $z^m_s=\{z_i\mid i\in\mathcal{H}\}$ the masked ones.
We employ a cross-attention regressor $\mathcal{R}_{\phi}$ that uses $z^v_s$ as keys and values while querying with learnable mask tokens plus positional embeddings, yielding reconstructed latent representations $\hat{z}^m_s=\mathcal{R}_{\phi}(z^v_s)$.

The reconstruction loss compares the predicted masked tokens with the expert-projected targets:
\begin{equation}
\mathcal{L}_{\mathrm{recon}}
= \frac{1}{|\mathcal{H}|}
  \sum_{i\in\mathcal{H}}
  \frac{1}{J}\sum_{j=1}^{J}
  \|\, \hat{z}^m_{s,i} - \tilde{z}^{(j)}_i \,\|_2^2 .
\label{eq:recon}
\end{equation}
Unlike pixel-level reconstruction, this formulation constrains the model to infer feature relationships.

\textbf{Edge-Aware Masking.}
To ensure that reconstruction focuses on informative regions,
we employ an edge-aware masking strategy assigning higher masking probability to patches with strong gradients,
typically corresponding to bone and joint boundaries.
Let $p_i$ denote the normalized edge strength of patch $i$.
The sampling distribution is:
\begin{equation}
q_i = (1-\alpha)\frac{p_i}{\sum_k p_k}
      + \alpha\frac{1}{N}, \quad
\mathcal{H} \sim \mathrm{SampleWithoutReplacement}(q, \lfloor rN \rfloor),
\label{eq:mask}
\end{equation}
where $r$ is the mask ratio and $\alpha\in[0,1]$ controls the trade-off between edge-based and random masking.
By emphasizing structurally salient regions, the model learns geometry-aware features and spatial continuity. The masked latent regression captures inter-bone and joint relationships, improving sensitivity to structural deformation and localization errors for segmentation, landmark detection, and joint-level scoring.

\subsection{EMA Consistency}
\label{sec:ema}

To stabilize optimization and avoid overfitting to instantaneous expert outputs,
we use an exponential moving average (EMA) of the student encoder as a smoothed teacher providing stable targets for self-consistency.
After each update step, parameters are updated as
\begin{equation}
\theta_m \gets \mu\, \theta_m + (1-\mu)\, \theta , \quad \mu = 0.996,
\label{eq:ema_update}
\end{equation}
where $\mu$ is a constant momentum coefficient controlling update smoothness.
Following standard self-distillation frameworks such as DINO and BYOL,
we empirically set $\mu = 0.996$ for stable evolution of the EMA teacher.
The EMA encoder produces latent features
$\bar{z}_s = f_{\theta_m}(x) = [\bar{z}_{\mathrm{cls}}; \bar{z}_1, \dots, \bar{z}_N]$.
We apply a consistency loss between the current and EMA representations:
\begin{equation}
\mathcal{L}_{\mathrm{ema}}
= \frac{1}{|\mathcal{V}|+|\mathcal{H}|}
  \sum_{i \in \mathcal{V} \cup \mathcal{H}}
  \|\, z_{s,i} - \bar{z}_{s,i} \,\|_2^2 .
\label{eq:ema_loss}
\end{equation}
This constraint regularizes feature evolution. The temporal averaging of encoder weights acts as a low-pass filter, preventing instability from stochastic expert signals or local minima. It encourages gradual refinement of high-level concepts and maintains manifold continuity. 

The overall pretraining objective combines alignment, reconstruction, and consistency terms:
\begin{equation}
\hspace{-0.5em}
\mathcal{L}_{\mathrm{total}}
= \mathcal{L}_{\mathrm{align}}
+ \lambda_{\mathrm{recon}} \mathcal{L}_{\mathrm{recon}}
+ \lambda_{\mathrm{ema}} \mathcal{L}_{\mathrm{ema}}
+ \lambda_{\mathrm{cls}} \mathcal{L}_{\mathrm{cls}} .
\label{eq:total}
\end{equation}

The full pretraining loop is summarized in~\Cref{alg:handxfm}, and implementation details are provided in Supp.~\ref{supp:pretrain_config}.

\begin{algorithm}[tb]
  \caption{HandXFM pretraining with multi-expert masked latent alignment}
  \label{alg:handxfm}
  \begin{algorithmic}
    \STATE {\bfseries Input:} Unlabeled hand X-ray dataset $\mathcal{D}$; student encoder $f_{\theta}$;
    experts $\{t_j\}_{j=1}^{J}$; regressor $\mathcal{R}_{\phi}$; projection heads $\{P_j\}$;
    EMA encoder $f_{\theta_m}$; loss weights $\lambda_{\mathrm{recon}}, \lambda_{\mathrm{ema}}, \lambda_{\mathrm{cls}}$.
    \STATE {\bfseries Output:} Student encoder $f_{\theta}$ and EMA teacher $f_{\theta_m}$.
    \STATE Initialize $\theta, \phi, \{P_j\}$; set $\theta_m \gets \theta$.
    \FOR{each minibatch $x \sim \mathcal{D}$}
      \STATE Sample binary mask $M$ using edge-aware PatchMasker (\cref{eq:mask}).
      \STATE Obtain visible index set $\mathcal{V}$ and masked index set $\mathcal{H}$.
      \STATE Compute student tokens $z_s = f_{\theta}(x) = [z_{\mathrm{cls}}; z_1,\dots,z_N]$.
      \FOR{each expert $t_j$}
        \STATE Compute expert tokens $z_{t_j} = t_j(x)$ and project $\tilde{z}^{(j)} = P_j z_{t_j}$.
      \ENDFOR
      \STATE Compute alignment loss $\mathcal{L}_{\mathrm{align}}$ (\cref{eq:align})
             and CLS loss $\mathcal{L}_{\mathrm{cls}}$ (\cref{eq:cls}).
      \STATE Predict masked tokens $\hat{z}_s^{m} = \mathcal{R}_{\phi}(z_s^{v})$
             and compute reconstruction loss $\mathcal{L}_{\mathrm{recon}}$ (\cref{eq:recon}).
      \STATE Compute EMA teacher tokens $\bar{z}_s = f_{\theta_m}(x)$
             and loss $\mathcal{L}_{\mathrm{ema}}$ (\cref{eq:ema_loss}).
    \STATE Compute total loss $\mathcal{L}_{\mathrm{total}}$ (\cref{eq:total}) and update $\theta, \phi, \{P_j\}$ with AdamW.
      \STATE Update EMA teacher: $\theta_m \gets \mu\,\theta_m + (1-\mu)\,\theta$.
    \ENDFOR
  \end{algorithmic}
\end{algorithm}

\section{Experiments}
\label{sec:experiment}
\subsection{Datasets}
\label{sec:datasets}

To study hand radiograph foundation modeling under realistic clinical heterogeneity, we assemble a large corpus spanning development, disease grading, abnormality screening, localization, and structural understanding. The datasets differ in patient population, annotation granularity, label units, and downstream objectives, reflecting the fragmented nature of real-world hand radiograph use. To our knowledge, this is the largest open-access benchmark collection in this domain. \textbf{Pretraining uses training splits only; all val/test splits and external-validation datasets are excluded}.


\begin{itemize}[leftmargin=*, itemsep=2pt, topsep=0pt, parsep=0pt, partopsep=0pt]

\item \textbf{Bone Age Prediction.}
We evaluate bone age estimation on three datasets:
\textbf{RSNA}~\cite{Halabi2019RSNA} (14,036 radiographs; months),
\textbf{RHPE}~\cite{gonzalez2020simba} (6,279 images; months + sex),
and \textbf{Digital Hand Atlas}~\cite{dha} (1,103 images; years + sex), used for external validation.

\item \textbf{SvH Score Regression.}
To assess rheumatoid arthritis severity, we use
\textbf{AIRINIIReumHands}~\cite{Stolpovsky_2023_ICCV} (330 radiographs; joint-level erosion and JSN annotations)
and the \textbf{Anhui} dataset~\cite{anhui} (3,818 bilateral radiographs with image-level averaged scores).

\item \textbf{Abnormality Classification.}
We use the MURA hand subset~\cite{rajpurkar2018mura}
(6,003 radiographs labeled normal/abnormal),
restricting to the hand category.

\item \textbf{Closed-end VQA.}
We use 108 hand X-rays from \textbf{MedPix}~\cite{medpix}, treating captions as questions and titles as answers.

\item \textbf{Joint Detection and Bone Segmentation (New Annotations).}
We curate two RSNA-based datasets:
a joint detection set (6,130 images with 17 landmark boxes; \Cref{tab:joint_definitions})
and a bone segmentation set (549 images with 19-class masks; \Cref{tab:bone_definitions}).
Annotations follow \emph{Terminologia Anatomica}~\cite{terminologia_anatomica} and are aligned with \emph{RSNA RadLex}~\cite{radlex}.

\end{itemize}

\paragraph{Data separation and baseline protocols.}
Only the designated training partitions contribute
images to unlabeled HandXFM pretraining; validation,
test, and external-evaluation images are excluded.
For RSNA-derived localization benchmarks, detector
and segmenter training use RSNA training images,
while evaluation uses the disjoint validation
partition. Unless otherwise specified, pretrained
visual baselines are fully fine-tuned using the
same downstream partitions and task-specific
evaluation protocols. Detection and segmentation
backbones share the same DETR-style downstream
architecture, whereas results adopted from
published task-specific systems are reported
separately. Detailed dataset allocation and
overlap controls are provided in
\Cref{tab:dataset_allocation}.

\subsection{Downstream Tasks}
\label{sec:downstream_tasks}

\begin{table}[t]
\centering
\caption{Regression results on SvH scoring and bone age prediction.}
\label{tab:regression_main}
\small
\setlength{\tabcolsep}{2.5pt}
\renewcommand{\arraystretch}{0.95}
\begin{tabular}{@{}lcccc!{\hspace{0.6em}\vrule width 0.4pt\hspace{0.6em}}lcccc@{}}
\toprule
\multicolumn{5}{c!{\hspace{0.6em}\vrule width 0.4pt\hspace{0.6em}}}{\textbf{SvH Scoring}} &
\multicolumn{5}{c}{\textbf{Bone Age Prediction}} \\
\cmidrule(lr){1-5}\cmidrule(lr){6-10}

\multicolumn{5}{c!{\hspace{0.6em}\vrule width 0.4pt\hspace{0.6em}}}{\textbf{AIRINIIReumHands (Erosion)}} &
\multicolumn{5}{c}{\textbf{RSNA (Monthly Labels)}} \\
Method & RMSE$\downarrow$ & MAE$\downarrow$ & $R^2\uparrow$ & PCC$\uparrow$ &
Method & RMSE$\downarrow$ & MAE$\downarrow$ & $R^2\uparrow$ & PCC$\uparrow$ \\
\midrule
RheumaVIT~\cite{Stolpovsky_2023_ICCV} & 0.445 & 0.360 & 0.373 & 0.766 &
CCR-BAA~\cite{ccrBAA} & 11.911 & 8.873 & 0.918 & 0.959 \\
DINOv3~\cite{siméoni2025dinov3} & 0.525 & 0.300 & 0.127 & 0.426 &
AE-BoNet~\cite{AEbonet} & 13.799 & 10.190 & 0.891 & 0.944 \\
RadFM~\cite{wu2025_radfm} & 0.409 & 0.231 & 0.403 & 0.797 &
DINOv3~\cite{siméoni2025dinov3} & 12.702 & 9.106 & 0.907 & 0.953 \\
RadCLIP~\cite{radclip} & 0.402 & 0.136 & 0.487 & 0.739 &
RadFM~\cite{wu2025_radfm} & 18.395 & 13.752 & 0.805 & 0.898 \\
HandXFM & \textbf{0.298} & \textbf{0.111} & \textbf{0.719} & \textbf{0.929} &
RadCLIP~\cite{radclip} & 12.773 & 9.379 & 0.906 & 0.954 \\
 &  &  &  &  &
HandXFM & \textbf{9.932} & \textbf{7.343} & \textbf{0.943} & \textbf{0.972} \\
\midrule

\multicolumn{5}{c!{\hspace{0.6em}\vrule width 0.4pt\hspace{0.6em}}}{\textbf{AIRINIIReumHands (JSN)}} &
\multicolumn{5}{c}{\textbf{RHPE (Monthly Labels)}} \\
Method & RMSE$\downarrow$ & MAE$\downarrow$ & $R^2\uparrow$ & PCC$\uparrow$ &
Method & RMSE$\downarrow$ & MAE$\downarrow$ & $R^2\uparrow$ & PCC$\uparrow$ \\
\midrule
RheumaVIT~\cite{Stolpovsky_2023_ICCV} & 0.758 & 0.601 & 0.538 & 0.847 &
ConvNeXt~\cite{rhpeconvnext} & 14.288 & 11.497 & 0.877 & 0.965 \\
DINOv3~\cite{siméoni2025dinov3} & 0.714 & \textbf{0.558} & 0.591 & 0.803 &
BoNet~\cite{bonet} & 11.246 & 8.708 & 0.926 & 0.962 \\
RadFM~\cite{wu2025_radfm} & 0.792 & 0.618 & 0.496 & 0.740 &
DINOv3~\cite{siméoni2025dinov3} & 27.828 & 21.856 & 0.550 & 0.790 \\
RadCLIP~\cite{radclip} & 0.792 & 0.635 & 0.497 & 0.861 &
RadFM~\cite{wu2025_radfm} & 23.887 & 18.233 & 0.668 & 0.820 \\
HandXFM & \textbf{0.705} & 0.608 & \textbf{0.601} & \textbf{0.885} &
RadCLIP~\cite{radclip} & 12.320 & 9.656 & 0.912 & 0.955 \\
 &  &  &  &  &
HandXFM & \textbf{11.073} & \textbf{8.363} & \textbf{0.929} & \textbf{0.966} \\
\midrule

\multicolumn{5}{c!{\hspace{0.6em}\vrule width 0.4pt\hspace{0.6em}}}{\textbf{Anhui (Whole-hands Image)}} &
\multicolumn{5}{c}{\textbf{DHA (External Validation; Year Labels)}} \\
Method & RMSE$\downarrow$ & MAE$\downarrow$ & $R^2\uparrow$ & PCC$\uparrow$ &
Method & RMSE$\downarrow$ & MAE$\downarrow$ & $R^2\uparrow$ & PCC$\uparrow$ \\
\midrule
ResNet dw-50~\cite{anhui} & 24.150 & 16.688 & 0.669 & 0.835 &
CCR-BAA~\cite{ccrBAA} & 1.528 & 1.193 & 0.899 & 0.952 \\
ResNet-50~\cite{anhui_baseline2} & 30.967 & 22.185 & 0.456 & 0.678 &
AE-BoNet~\cite{AEbonet} & 1.636 & 1.233 & 0.884 & 0.941 \\
DINOv3~\cite{siméoni2025dinov3} & 31.232 & 20.301 & 0.447 & 0.702 &
DINOv3~\cite{siméoni2025dinov3} & 1.909 & 1.418 & 0.842 & 0.930 \\
RadFM~\cite{wu2025_radfm} & 30.513 & 20.595 & 0.472 & 0.721 &
RadFM~\cite{wu2025_radfm} & 1.799 & 1.365 & 0.859 & 0.929 \\
RadCLIP~\cite{radclip} & 22.671 & 14.642 & 0.709 & 0.869 &
RadCLIP~\cite{radclip} & 1.864 & 1.420 & 0.849 & 0.931 \\
HandXFM & \textbf{18.231} & \textbf{12.356} & \textbf{0.812} & \textbf{0.903} &
HandXFM & \textbf{1.269} & \textbf{0.990} & \textbf{0.930} & \textbf{0.965} \\
\bottomrule
\end{tabular}
\vskip -0.12in
\end{table}

To demonstrate HandXFM's versatility, we apply it as an encoder backbone for diverse downstream tasks (regression, classification, segmentation, detection, and VQA) using lightweight task-specific heads. These heads are intentionally small, ensuring performance reflects representation quality rather than head capacity. The following subsections describe each task and its protocol.

\subsubsection{SvH Scoring}

We evaluate HandXFM on two rheumatoid arthritis datasets with different supervision levels: AIRINIIReumHands~\cite{Stolpovsky_2023_ICCV} (joint-level) and Anhui~\cite{anhui} (image-level). Because of this granularity difference, we design two lightweight regression heads built on the same pretrained HandXFM encoder.

\begin{itemize}[leftmargin=*, itemsep=2pt, topsep=2pt, parsep=0pt, partopsep=0pt]

\item \textbf{Joint-level head.}
For AIRINIIReumHands, we attach two parallel regression branches to predict erosion and JSN scores from the pooled representation.
Missing joint annotations are handled with a masked Smooth-$\ell_1$ loss so each image contributes supervision only where labels exist.

\item \textbf{Whole-hands head.}
For the Anhui dataset, we use a shallow regression head on the CLS token to predict the global SvH score of both hands.

\end{itemize}

We report RMSE, MAE, $R^2$, and PCC,
which together assess absolute regression error, explained variance, and agreement with radiologist scoring.
The results are summarized in ~\Cref{tab:regression_main}. HandXFM shows consistent improvements across both joint-level and whole-hand SvH scoring tasks.

On AIRINIIReumHands, HandXFM achieves PCCs of 0.929 (erosion) and 0.885 (JSN), outperforming all baselines. Although some baselines achieve lower JSN MAE, HandXFM obtains the highest PCC among the evaluated methods, indicating stronger correlation with the reference damage scores. We report a multi-seed stability analysis for it in Supp.~\ref{supp:appendix_multiseed_airinii}, given its relatively small dataset size.

On the Anhui dataset (single score for both hands), HandXFM reduces RMSE from 24–31 to 18.2 and improves $R^2$ to 0.812, surpassing RadCLIP (22.67, 0.709) and RadFM (30.51, 0.472). These gains suggest a stronger representation that aggregates bilateral lesions into a global disease estimate.


\subsubsection{Bone Age Prediction}
\label{sec:exp_boneage}
We assess bone age estimation, a benchmark for maturation, on three datasets: RSNA, RHPE, and Digital Hand Atlas (DHA) for external validation. Acknowledging sexual dimorphism, regression heads jointly encode the image and a sex variable, which is projected and concatenated with the image embedding. All models use a Smooth-$\ell_1$ loss.

\begin{itemize}[leftmargin=*, itemsep=2pt, topsep=2pt, parsep=0pt, partopsep=0pt]

\item \textbf{RSNA.}
As shown in~\Cref{tab:regression_main}, HandXFM achieves the lowest MAE (7.34 months) and highest PCC (0.972), outperforming hand-specific pipelines (AE-BoNet, CCR-BAA) and foundation-model baselines (RadFM, RadCLIP). Strong $R^2$ and PCC indicate preserved maturity ordering beyond absolute error, suggesting multi-expert pretraining captures subtle developmental cues.

\item \textbf{RHPE.}
HandXFM achieves the best overall performance on RHPE, reducing MAE to 8.36 months while reaching $R^2=0.929$ and PCC=0.966. It outperforms DINOv3, RadFM, RadCLIP, ConvNeXt, and BoNet, indicating that the pretrained representation effectively captures bone structure and epiphyseal development for pediatric bone age estimation.

\item \textbf{External validation on DHA.}
Training on RSNA (months) and evaluating on DHA (years), HandXFM maintains high correlation (PCC=0.965, $R^2=0.930$). As the head is not tuned on DHA, this reflects strong backbone generalization and stable anatomical representations.

\end{itemize}

\subsubsection{Abnormality Classification}
\label{sec:exp_mura}

We evaluate abnormality classification on the MURA hand subset and summarize the results in \Cref{tab:mura_main}. HandXFM serves as the visual encoder, followed by a CLS-only head with two linear layers applied to the encoder's CLS token. We train with BCE-with-logits and report AUC, AUPRC, ACC, F1, precision (P), and recall (R), using a weighted sampler to address class imbalance. Baselines include a self-taught semi-supervised anomaly detection method (ST-SSAD)~\cite{st_ssad}, a supervised DenseNet-169 pipeline~\cite{mura_d169}, and radiology foundation models (RadFM~\cite{wu2025_radfm}, RadCLIP~\cite{radclip}).

\begin{table}[t]
\caption{MURA XR\_HAND classification (normal vs.\ abnormal).}
\label{tab:mura_main}
\centering
\small
\setlength{\tabcolsep}{4pt}
\begin{tabular*}{\linewidth}{@{\extracolsep{\fill}}lcccccc@{}}
\toprule
Method & AUC$\uparrow$ & AUPRC$\uparrow$ & ACC$\uparrow$ & F1$\uparrow$ & P$\uparrow$ & R$\uparrow$ \\
\midrule
ST-SSAD~\cite{st_ssad} & 0.5966 & 0.4969 & 0.5978 & 0.0887 & 0.6429 & 0.0476 \\
DenseNet-169~\cite{mura_d169} & 0.8234 & 0.7788 & 0.7109 & 0.6826 & 0.6217 & 0.7566 \\
DINOv3~\cite{siméoni2025dinov3} & 0.7405 & 0.6667 & 0.6741 & 0.6589 & 0.5900 & 0.7460 \\
RadFM~\cite{wu2025_radfm} & 0.6114 & 0.5214 & 0.4891 & 0.5983 & 0.4419 & \textbf{0.9259} \\
RadCLIP~\cite{radclip} & 0.8460 & 0.8084 & 0.7500 & 0.6000 & \textbf{0.9231} & 0.4444 \\
HandXFM (ours) & \textbf{0.8599} & \textbf{0.8213} & \textbf{0.7857} & \textbf{0.6903} & 0.8843 & 0.5661 \\
\bottomrule
\end{tabular*}
\vskip -0.25in
\end{table}

HandXFM achieves the best performance on the MURA hand subset. Compared with the DenseNet-169 pipeline, it improves AUC from 0.823 to 0.860, AUPRC from 0.779 to 0.821, and accuracy from 71.1\% to 78.6\%, while also achieving the highest F1 (0.690). With a CLS-only head, HandXFM maintains high precision (P = 0.884) and competitive recall (R = 0.566), yielding a strong precision--recall trade-off. In contrast, RadCLIP favors precision (P = 0.923; R = 0.444) and RadFM favors recall (R = 0.926; P = 0.442), leading to weaker discrimination. Relative to DINOv3-only, multi-expert pretraining adds anatomical priors and improves robustness to heterogeneous image quality.

\subsubsection{Joint Detection}
\label{sec:exp_detection}

We evaluate landmark-level joint detection on a 17-class dataset covering DIP, PIP, MCP joints, and the radius, ulna, and wrist center.
A DETR-style decoder~\cite{10.1007/978-3-030-58452-8_13,huang2024deim,huang2025deimv2} is attached to the HandXFM encoder: patch tokens are projected to 256-D embeddings with learnable 2D positional encodings and decoded by a 6-layer Transformer with 100 queries.
The decoder predicts class logits and normalized boxes via parallel MLP heads, and is trained with Hungarian matching using a composite loss (classification, $\ell_1$, and GIoU) optimized by AdamW with a cosine schedule.
Evaluation follows COCO-style mAP metrics (mAP@0.5:0.95, mAP@0.5, mAP@0.75).

As shown in \Cref{tab:det_seg_main}, HandXFM outperforms all matched DETR-based baselines, including ResNet50, ViT, ConvNeXt, native DINOv3, and domain-adapted DINOv3, and also surpasses the supervised YOLOv12~\cite{tian2025yolov12} in mAP@0.5:0.95. It achieves the best high-IoU localization quality (mAP@0.75 = 0.9026), indicating more accurate landmark localization. These results suggest that the gain is attributable to the HandXFM backbone rather than decoder choice alone. Its attention-guided representations capture bone and joint topology, enabling geometric reasoning over part--whole relationships, as illustrated in the left panel of \Cref{fig:qual_vis}, with qualitative examples in Supp.~\ref{supp:resuls_details}.

\subsubsection{Bone Segmentation}
\label{sec:exp_segmentation}

We perform bone segmentation on a 19-class dataset covering phalanges and metacarpals. Following the detection framework, we adopt a DETR architecture: the HandXFM encoder outputs patch tokens projected to a 256-dimensional feature map and processed by a transformer encoder--decoder. Each decoder query predicts boxes and class labels, while a lightweight head refines them into masks. The head upsamples 14$\times$14 feature maps to 224$\times$224 via bilinear interpolation and applies two convolutional layers with group normalization. Training jointly optimizes a set loss (classification, $\ell_1$ box, and GIoU) and a segmentation loss (weighted cross-entropy + $0.5\times$ Dice). We report precision/recall and segmentation metrics (mAP@0.5:0.95, mAP@0.5, mAP@0.75).

As summarized in \Cref{tab:det_seg_main} and visualized in the left panel of \Cref{fig:qual_vis}, HandXFM achieves the best segmentation performance, outperforming compared backbones and YOLOv12, with mAP@0.5:0.95 improving to 0.6101 and precision rising to 0.9481. These gains are obtained without pixel pretraining supervision and under a matched downstream head, indicating that the improvement comes from the HandXFM backbone. The encoder’s latent representations, learned via multi-expert alignment and edge-aware masking, capture bone contours and spatial separations as structural cues, enabling the decoder to delineate complex boundaries with greater geometric fidelity. This suggests that HandXFM internalizes spatial structure during pretraining rather than relying on task-specific supervision.

\begin{table}[t]
\caption{Joint detection and bone segmentation results.}
\label{tab:det_seg_main}
\centering
\small
\setlength{\tabcolsep}{3pt}
\renewcommand{\arraystretch}{0.97}

\begin{minipage}[t]{0.49\linewidth}
\centering
\textbf{Joint detection (17-class)}\\
\resizebox{\linewidth}{!}{
\begin{tabular}{@{}lccc@{}}
\toprule
Method & mAP$_{50:95}\uparrow$ & mAP$_{50}\uparrow$ & mAP$_{75}\uparrow$ \\
\midrule
ResNet50 + DETR & 0.5303 & 0.9433 & 0.5409 \\
ViT + DETR & 0.5617 & 0.9508 & 0.6057 \\
ConvNeXt + DETR & 0.5517 & 0.9468 & 0.6023 \\
DINOv3 + DETR (native) & 0.5857 & 0.9630 & 0.6512 \\
YOLOv12 & 0.7391 & \textbf{0.9903} & 0.8752 \\
DINOv3 + DETR (adapted) & 0.5900 & 0.9650 & 0.6487 \\
HandXFM & \textbf{0.7636} & 0.9802 & \textbf{0.9026} \\
\bottomrule
\end{tabular}}
\end{minipage}
\hfill
\begin{minipage}[t]{0.49\linewidth}
\centering
\textbf{Bone segmentation (19-class)}\\
\resizebox{\linewidth}{!}{
\begin{tabular}{@{}lccc@{}}
\toprule
Method & P$\uparrow$ & R$\uparrow$ & mAP$_{50:95}\uparrow$ \\
\midrule
ResNet50 + DETR & 0.9426 & 0.9136 & 0.4191 \\
ViT + DETR & 0.9437 & 0.9073 & 0.3482 \\
ConvNeXt + DETR & 0.9454 & 0.9173 & 0.4356 \\
DINOv3 + DETR (native) & 0.9455 & 0.9191 & 0.4342 \\
YOLOv12 & 0.9006 & 0.9080 & 0.5051 \\
HandXFM & \textbf{0.9481} & \textbf{0.9264} & \textbf{0.6101} \\
\bottomrule
\end{tabular}}
\end{minipage}

\vskip -0.2in
\end{table}

\paragraph{Comparison with semantic segmentation baselines.}
We additionally compare HandXFM against U-Net
and nnU-Net using pixel-level segmentation
metrics on the same evaluation set
(\Cref{tab:supp_semantic_seg_baselines}).
HandXFM achieves the highest foreground
precision (0.4596), recall (0.4984), and
F1 score (0.4777), while maintaining pixel
accuracy comparable to U-Net and nnU-Net.
These measurements complement the COCO-style
instance segmentation results, which evaluate
the localization and separation of individual
anatomical bone instances rather than only
pixel-level foreground prediction.

\subsubsection{Visual Question Answering}
\label{sec:exp_vqa}

To evaluate cross-modal capability, we conduct a closed-ended visual question answering (VQA) task on the hand subset of MedPix (108 image--question pairs)~\cite{Siragusa_medpix_2025}. We use a fixed train/test split, and all test images are excluded from the unlabeled pretraining corpus. Each sample pairs a radiograph with a caption-derived query, using the diagnostic title as the ground-truth answer. HandXFM combines its visual encoder with a frozen BiomedCLIP text encoder via cross-attention, followed by a two-layer MLP classifier. Models are trained with cross-entropy and evaluated using Top-1 and Top-5 accuracy. HandXFM achieves 0.6364 Top-1 and 0.9091 Top-5 accuracy, matching RadCLIP in this small-scale setting. Given the limited size of the MedPix hand subset, we interpret this result as evidence of cross-modal capability rather than as a definitive ranking between models, especially since HandXFM uses a visual backbone pretrained without paired image--text supervision.

\begin{figure}[t]
\centering
\begin{minipage}[t]{0.575\linewidth}
    \centering
    \includegraphics[width=\linewidth]{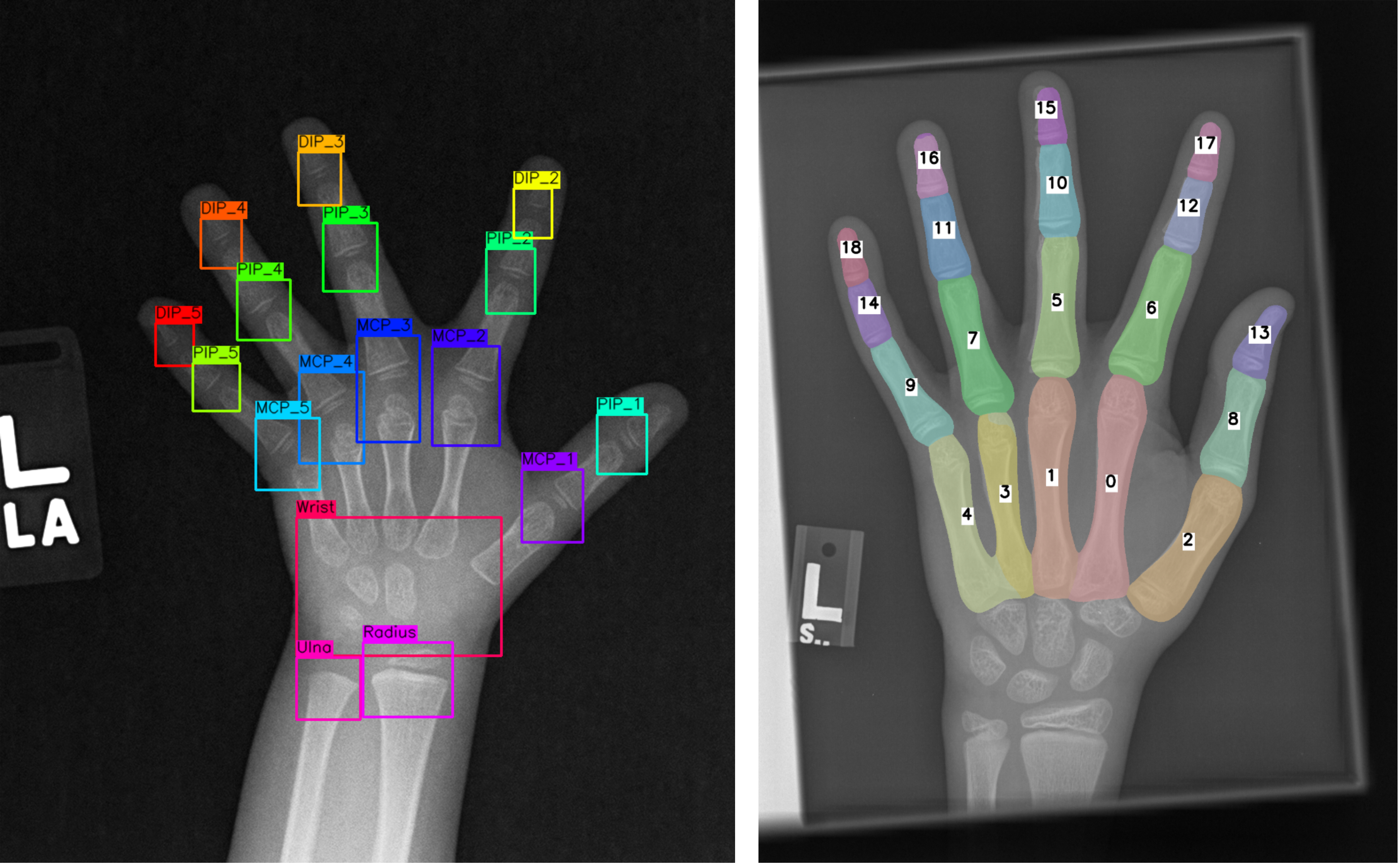}
\end{minipage}
\hfill
\begin{minipage}[t]{0.385\linewidth}
    \centering
    \includegraphics[width=\linewidth]{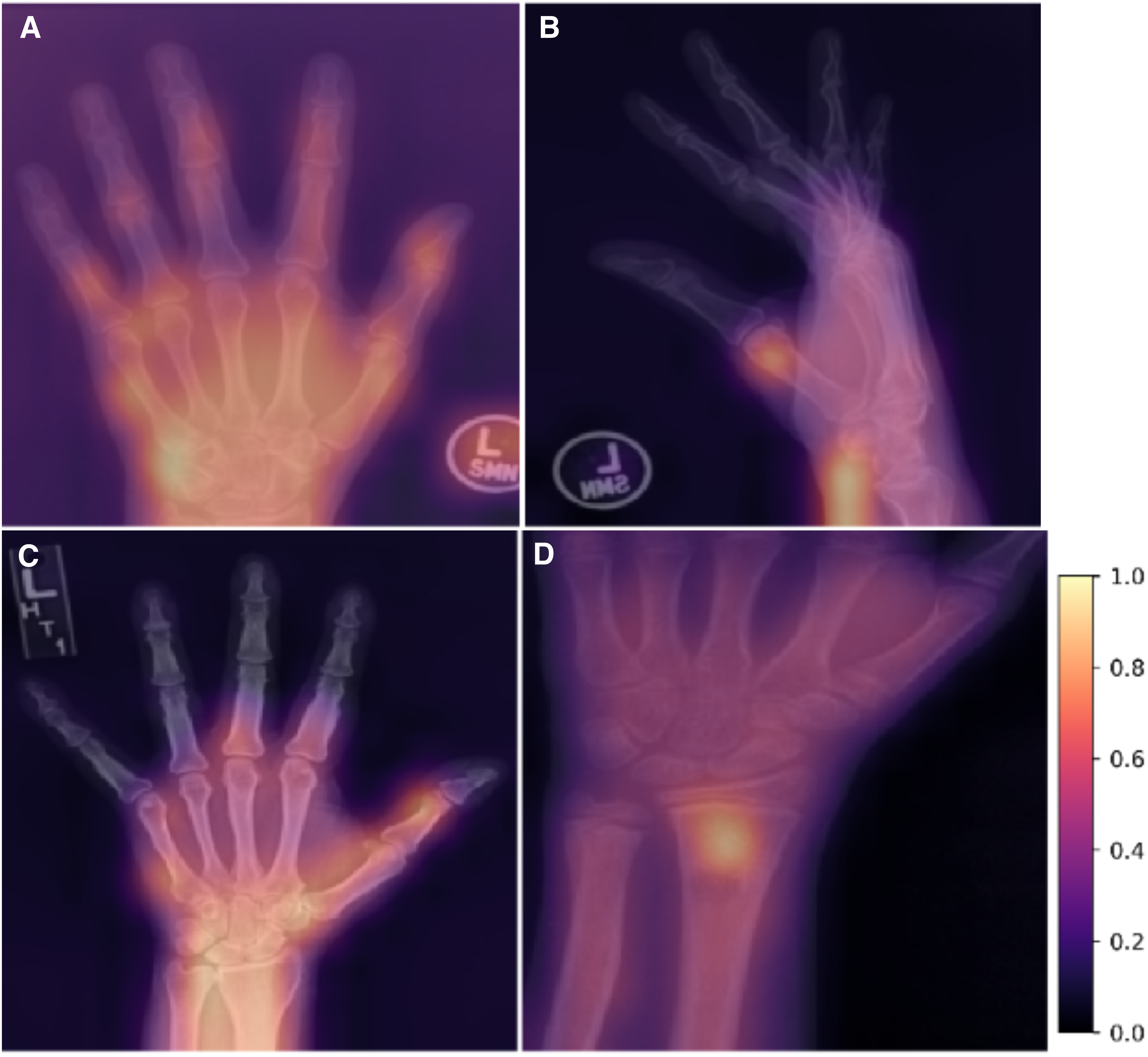}
\end{minipage}
\caption{
\textbf{Qualitative visualizations.}
Left: predicted examples of joint detection and bone segmentation, where bounding boxes denote localized joints and colored masks represent segmented bones.
Right: text--image cross-attention maps for clinical terms: (A) joint, (B) scaphotrapezial, (C) metacarpals, and (D) radius.
}
\label{fig:qual_vis}
\vskip -0.2in
\end{figure}

\textbf{Visualization.}
As shown in the right panel of \Cref{fig:qual_vis}, cross-attention maps illustrate clinical terms grounded in radiographic regions. Examples include “joint,” “scaphotrapezial,” “metacarpals,” and “radius.” The maps show correspondence between tokens and anatomical structures, demonstrating effective alignment between clinical concepts and spatial regions, and supporting interpretability.

\subsection{Ablation Study}
\label{sec:ablation}
We conduct a series of ablation studies to analyze the contributions of design choices in HandXFM.
Specifically, we examine: (1) the number of expert teachers used during pretraining,
(2) the role of the alignment and reconstruction objectives,
and (3) the impact of edge-aware masking.

\subsubsection{Teacher Ensemble Ablation}
\label{sec:ablation_experts}

We compare three pretraining configurations: 
(i) \emph{No teacher} (DINOv3 student only), 
(ii) \emph{Single teacher} (BiomedCLIP or Chest X-ray ViT), and 
(iii) \emph{Dual teachers} (HandXFM).

\begin{itemize}[leftmargin=*, itemsep=2pt, topsep=2pt, parsep=0pt, partopsep=0pt]

\item \textbf{Dual-teacher distillation is strongest.}
On bone age (\Cref{tab:supp_ablation_expert_reg}), HandXFM reduces MAE from 9.11 (DINOv3-only) to 7.34 and improves PCC from 0.953 to 0.972; 
on joint detection, mAP@0.5:0.95 increases from 0.59 to 0.76.
On MURA (\Cref{tab:supp_ablation_expert_mura}), AUC rises from 0.74 to 0.86.

\item \textbf{Single teachers show task-aligned strengths.}
The Chest teacher performs best on structure-sensitive targets; for AIRINIIReumHands JSN it achieves RMSE 0.591 and MAE 0.380. BiomedCLIP improves semantic cues but is weaker on fine-grained structural scoring. Combining both yields the most balanced representation, achieving best or near-best performance across bone age, joint detection (\Cref{tab:supp_ablation_expert_joint}), and MURA classification.

\end{itemize}

\subsubsection{Ablation on Loss Composition}
\label{sec:ablation_loss}

We compare three settings: (i) \emph{Alignment-only} (teacher–student alignment), (ii) \emph{Reconstruction-only} (masked token prediction), and (iii) \emph{Combined}. 
Results are summarized in \Crefrange{tab:supp_ablation_loss_reg}{tab:supp_ablation_loss_bone}.

\begin{itemize}[leftmargin=*, itemsep=2pt, topsep=2pt, parsep=0pt, partopsep=0pt]

\item \textbf{Combined losses are most reliable.}
They achieve best or near-best results across RSNA and RHPE bone age (lowest MAE, highest PCC), MURA (AUC/ACC), and joint detection (mAP@0.5:0.95).

\item \textbf{Alignment-only.}
It excels on some fine-grained regressions (e.g., AIRINIIReumHands erosion MAE) but underperforms on JSN, indicating limited spatial coverage.

\item \textbf{Reconstruction-only.}
It remains competitive, trailing the combined setup but surpassing alignment-only on JSN and detection, showing masked prediction improves spatial completeness.

\end{itemize}

Overall, alignment sharpens teacher-consistent semantics, reconstruction enhances localization, and combining them yields the most robust cross-task performance.

\subsubsection{Ablation on Masking Strategy}
\label{sec:ablation_mask}


We compare the edge-aware masking scheme with a standard random masking baseline during pretraining. Both configurations use a 50\% masking ratio and identical training settings. They differ only in how masked patch positions are sampled. Edge-aware masking assigns higher sampling probabilities to patches with strong image gradients, using an edge--uniform mixture with $\alpha=0.5$.

Across most regression (\Cref{tab:supp_ablation_mask_reg}) and classification (\Cref{tab:supp_ablation_mask_mura}) tasks, edge-aware masking consistently outperforms random masking, indicating that emphasizing structurally salient regions benefits both semantic and quantitative prediction. 
The advantage is more pronounced for localization-heavy tasks. 
For joint detection (\Cref{tab:supp_ablation_mask_joint}), mAP@0.5:0.95 improves from 0.7125 to 0.7636 and mAP@0.75 from 0.8485 to 0.9026; for bone segmentation (\Cref{tab:supp_ablation_mask_bone}), instance accuracy increases from 0.9713 to 0.9751 and mAP@0.5:0.95 from 0.4680 to 0.6101.

Overall, edge-aware masking provides stronger
localization performance, while random masking
remains competitive on selected regression
and pixel-level segmentation metrics.

\subsubsection{Additional Representation Analysis}
\label{sec:representation_analysis}

We complement the component ablations with additional experiments examining matched in-domain pretraining and frozen-backbone representation transfer.

\paragraph{Matched pretraining control.}
To distinguish the effect of the HandXFM pretraining objective from additional exposure to hand radiographs, we compare native DINOv3, DINOv3 adapted on the same hand-radiograph training corpus, and HandXFM using the same joint-detection protocol. As shown in \Cref{tab:det_seg_main}, the domain-adapted DINOv3 achieves mAP@0.5:0.95 of 0.5900, compared with 0.5857 for native DINOv3 and 0.7636 for HandXFM. The improvement obtained by HandXFM suggests that its pretraining formulation contributes beyond additional in-domain data exposure.

\paragraph{Frozen-backbone evaluation.}
We also evaluate the pretrained encoders
without updating their backbone parameters.
On Anhui whole-hand SvH scoring, frozen
HandXFM achieves PCC 0.684 and
$R^2=0.333$, compared with PCC 0.654
and $R^2=0.241$ for RadCLIP
(\Cref{tab:anhui_frozen_backbone}).
These results indicate that the HandXFM
representation retains task-relevant
information accessible to a lightweight
regression head, even without end-to-end
backbone fine-tuning.

\section{Conclusion}
\label{sec:5_conclusion}
We presented HandXFM, a multi-expert foundation model for comprehensive hand X-ray understanding.
By aligning a DINOv3 student with two complementary teachers, BiomedCLIP for semantic priors and a Chest X-ray ViT for structural cues, HandXFM learns transferable representations that generalize across diverse downstream tasks, including bone-age and SvH regression, abnormality classification, joint detection, bone segmentation, and visual question answering.
Extensive experiments show consistent improvements over prior baselines and strong generalization to unseen datasets such as the Digital Hand Atlas.
Ablation studies highlight the effectiveness of multi-expert distillation, the synergy between alignment and reconstruction losses, and the benefit of edge-aware masking for structural localization.
Future work may extend HandXFM to integrate a broader spectrum of complementary expert knowledge beyond the current modalities.
Overall, HandXFM provides a unified and interpretable foundation for musculoskeletal imaging, illustrating how multi-expert distillation bridges structural precision with semantic understanding and offers a scalable blueprint for domain-specialized medical foundation models.

\begin{ack}
This work was supported in part by the Canada
Research Chairs Tier II Program (CRC-2021-00482),
the Canadian Institutes of Health Research
(PJT-212993), and the Canada Foundation for
Innovation (CFI) John R. Evans Leaders Fund
(JELF, project no.~43481).

\textbf{Competing interests.}
The authors declare no competing financial interests.

\end{ack}


\bibliography{neurips_2026}

@article{svh_score,
  title        = {How to read radiographs according to the Sharp/van der Heijde method},
  author       = {Van der Heijde, D. M. F. M.},
  journal      = {The Journal of Rheumatology},
  year         = {2000},
  month        = jan,
  volume       = {27},
  number       = {1},
  pages        = {261--263},
  pmid         = {10648051}
}

@InProceedings{He_2022_CVPR,
    author    = {He, Kaiming and Chen, Xinlei and Xie, Saining and Li, Yanghao and Doll\'ar, Piotr and Girshick, Ross},
    title     = {Masked Autoencoders Are Scalable Vision Learners},
    booktitle = {Proceedings of the IEEE/CVF Conference on Computer Vision and Pattern Recognition (CVPR)},
    month     = {June},
    year      = {2022},
    pages     = {16000-16009}
}

@article{paschali2025_fm,
  title={Foundation models in radiology: what, how, why, and why not},
  author={Paschali, Magdalini and Chen, Zhihong and Blankemeier, Louis and Varma, Maya and Youssef, Alaa and Bluethgen, Christian and Langlotz, Curtis and Gatidis, Sergios and Chaudhari, Akshay},
  journal={Radiology},
  volume={314},
  number={2},
  pages={e240597},
  year={2025},
  publisher={Radiological Society of North America}
}

@article{wu2025_radfm,
  author    = {Chaoyi Wu and Xiaoman Zhang and Ya Zhang and Hui Hui and Yanfeng Wang and Weidi Xie},
  title     = {Towards Generalist Foundation Model for Radiology by Leveraging Web-Scale 2D\&3D Medical Data},
  journal   = {Nature Communications},
  year      = {2025},
  volume    = {16},
  number    = {1},
  pages     = {7866},
  doi       = {10.1038/s41467-025-62385-7},
  url       = {https://doi.org/10.1038/s41467-025-62385-7},
  publisher = {Nature Publishing Group},
  address   = {United Kingdom},
  month     = aug
}

@article{andleeb2025,
  author    = {Ifrah Andleeb and Bilal Zahid Hussain and Julie Joncas and Soraya Barchi and Marjolaine Roy-Beaudry and Stefan Parent and Guy Grimard and Hubert Labelle and Luc Duong},
  title     = {Automatic Evaluation of Bone Age Using Hand Radiographs and Pancorporal Radiographs in Adolescent Idiopathic Scoliosis},
  journal   = {Diagnostics (Basel)},
  year      = {2025},
  volume    = {15},
  number    = {4},
  pages     = {452},
  doi       = {10.3390/diagnostics15040452},
  url       = {https://doi.org/10.3390/diagnostics15040452},
  publisher = {MDPI},
  address   = {Switzerland},
  month     = feb
}

@article{alzubaidi2024_shoulder,
  author    = {Laith Alzubaidi and Asma Salhi and Mohammed A. Fadhel and Jinshuai Bai and Freek Hollman and Kristine Italia and Roberto Pareyon and A. S. Albahri and Chun Ouyang and Jose Santamar{\'i}a and Kenneth Cutbush and Ashish Gupta and Amin Abbosh and Yuantong Gu},
  title     = {Trustworthy Deep Learning Framework for the Detection of Abnormalities in X-Ray Shoulder Images},
  journal   = {PLOS ONE},
  year      = {2024},
  volume    = {19},
  number    = {3},
  pages     = {e0299545},
  doi       = {10.1371/journal.pone.0299545},
  pmid      = {38466693},
  pmc       = {PMC10927121},
  publisher = {Public Library of Science},
  address   = {United States},
  month     = mar
}

@article{sun2025_vlmreview,
  author    = {Sun, Yiyao and Wen, Xinran and Zhang, Yan and Jin, Lijun and Yang, Chunna and Zhang, Qianhui and Jiang, Mingchen and Xu, Zhaoyang and Guo, Wei and Su, Juan and Jiang, Xiran},
  title     = {Visual-Language Foundation Models in Medical Imaging: A Systematic Review and Meta-Analysis of Diagnostic and Analytical Applications},
  journal   = {Computer Methods and Programs in Biomedicine},
  year      = {2025},
  volume    = {268},
  pages     = {108870},
  doi       = {10.1016/j.cmpb.2025.108870},
  url       = {https://www.sciencedirect.com/science/article/pii/S0169260725002871},
  issn      = {0169-2607},
  publisher = {Elsevier},
  address   = {Netherlands},
  month     = aug
}

@article{bluethgen2025_chestvlm,
  author    = {Bluethgen, Christian and Chambon, Pierre and Delbrouck, Jean-Benoit and van der Sluijs, Rogier and Połacin, Małgorzata and Zambrano Chaves, Juan Manuel and Abraham, Tanishq Mathew and Purohit, Shivanshu and Langlotz, Curtis P. and Chaudhari, Akshay S.},
  title     = {A Vision--Language Foundation Model for the Generation of Realistic Chest X-Ray Images},
  journal   = {Nature Biomedical Engineering},
  year      = {2025},
  volume    = {9},
  number    = {4},
  pages     = {494--506},
  doi       = {10.1038/s41551-024-01246-y},
  url       = {https://doi.org/10.1038/s41551-024-01246-y},
  issn      = {2157-846X},
  publisher = {Nature Publishing Group},
  address   = {United Kingdom},
  month     = apr
}

@InProceedings{radford2021learning,
  title = 	 {Learning Transferable Visual Models From Natural Language Supervision},
  author =       {Radford, Alec and Kim, Jong Wook and Hallacy, Chris and Ramesh, Aditya and Goh, Gabriel and Agarwal, Sandhini and Sastry, Girish and Askell, Amanda and Mishkin, Pamela and Clark, Jack and Krueger, Gretchen and Sutskever, Ilya},
  booktitle = 	 {Proceedings of the 38th International Conference on Machine Learning},
  pages = 	 {8748--8763},
  year = 	 {2021},
  editor = 	 {Meila, Marina and Zhang, Tong},
  volume = 	 {139},
  series = 	 {Proceedings of Machine Learning Research},
  month = 	 {18--24 Jul},
  publisher =    {PMLR},
  url = 	 {https://proceedings.mlr.press/v139/radford21a.html},
}

@article{EVA,
  title={EVA: Exploring the Limits of Masked Visual Representation Learning at Scale},
  author={Fang, Yuxin and Wang, Wen and Xie, Binhui and Sun, Quan and Wu, Ledell and Wang, Xinggang and Huang, Tiejun and Wang, Xinlong and Cao, Yue},
  journal={arXiv preprint arXiv:2211.07636},
  year={2022}
}

@article{zhang2024biomedclip,
  title={A Multimodal Biomedical Foundation Model Trained from Fifteen Million Image–Text Pairs},
  author={Sheng Zhang and Yanbo Xu and Naoto Usuyama and Hanwen Xu and Jaspreet Bagga and Robert Tinn and Sam Preston and Rajesh Rao and Mu Wei and Naveen Valluri and Cliff Wong and Andrea Tupini and Yu Wang and Matt Mazzola and Swadheen Shukla and Lars Liden and Jianfeng Gao and Angela Crabtree and Brian Piening and Carlo Bifulco and Matthew P. Lungren and Tristan Naumann and Sheng Wang and Hoifung Poon},
  journal={NEJM AI},
  year={2024},
  volume={2},
  number={1},
  doi={10.1056/AIoa2400640},
  url={https://ai.nejm.org/doi/full/10.1056/AIoa2400640}
}

@inproceedings{wang2022medclip,
  title     = {MedCLIP: Contrastive Learning from Unpaired Medical Images and Text},
  author    = {Wang, Zifeng and Wu, Zhenbang and Agarwal, Dinesh and Sun, Jimeng},
  booktitle = {Proceedings of the Conference on Empirical Methods in Natural Language Processing (EMNLP)},
  year      = {2022},
  pages     = {3876--3887},
  doi       = {10.18653/v1/2022.emnlp-main.256},
  publisher = {Association for Computational Linguistics},
  address   = {United States},
  url       = {https://pubmed.ncbi.nlm.nih.gov/39144675/}
}

@InProceedings{chen2022m3ae,
    author="Chen, Zhihong
    and Du, Yuhao
    and Hu, Jinpeng
    and Liu, Yang
    and Li, Guanbin
    and Wan, Xiang
    and Chang, Tsung-Hui",
    editor="Wang, Linwei
    and Dou, Qi
    and Fletcher, P. Thomas
    and Speidel, Stefanie
    and Li, Shuo",
    title="Multi-modal Masked Autoencoders for Medical Vision-and-Language Pre-training",
    booktitle="Medical Image Computing and Computer Assisted Intervention -- MICCAI 2022",
    year="2022",
    publisher="Springer Nature Switzerland",
    address="Cham",
    pages="679--689",
    isbn="978-3-031-16443-9"
}

@inproceedings{gonzalez2020simba,
    title={SIMBA: Specific Identity Markers for Bone Age Assessment},
    author={Gonz{\'a}lez, Cristina and Escobar, Mar{\'\i}a and Daza, Laura and Torres, Felipe and Triana, Gustavo and Arbel{\'a}ez, Pablo},
    booktitle={International Conference on Medical Image Computing and Computer-Assisted Intervention},
    pages={753--763},
    year={2020},
    organization={Springer}
}

@article{Halabi2019RSNA,
  author    = {Halabi, Safwan S. and Prevedello, Luciano M. and Kalpathy-Cramer, Jayashree and
               Mamonov, Artem B. and Bilbily, Alexander and Cicero, Mark and Pan, Ian and
               Pereira, Lucas Ara{\'u}jo and Sousa, Rafael Teixeira and Abdala, Nitamar and
               Kitamura, Felipe Campos and Thodberg, Hans H. and Chen, Leon and Shih, George and
               Andriole, Katherine and Kohli, Marc D. and Erickson, Bradley J. and
               Flanders, Adam E.},
  title     = {The RSNA Pediatric Bone Age Machine Learning Challenge},
  journal   = {Radiology},
  year      = {2019},
  volume    = {290},
  number    = {2},
  pages     = {498--503},
  doi       = {10.1148/radiol.2018180736},
  pmid      = {30480490}
}

@ARTICLE{rajpurkar2018mura,
       author = {{Rajpurkar}, Pranav and {Irvin}, Jeremy and {Bagul}, Aarti and {Ding}, Daisy and {Duan}, Tony and {Mehta}, Hershel and {Yang}, Brandon and {Zhu}, Kaylie and {Laird}, Dillon and {Ball}, Robyn L. and {Langlotz}, Curtis and {Shpanskaya}, Katie and {Lungren}, Matthew P. and {Ng}, Andrew Y.},
        title = "{MURA: Large Dataset for Abnormality Detection in Musculoskeletal Radiographs}",
      journal = {arXiv e-prints},
         year = 2017,
        month = dec,
          eid = {arXiv:1712.06957},
        pages = {arXiv:1712.06957},
          doi = {10.48550/arXiv.1712.06957},
archivePrefix = {arXiv},
       eprint = {1712.06957},
 primaryClass = {physics.med-ph},
       adsurl = {https://ui.adsabs.harvard.edu/abs/2017arXiv171206957R}
}

@InProceedings{Stolpovsky_2023_ICCV,
    author    = {Stolpovsky, Alexander and Dakhova, Elizaveta and Druzhinina, Polina and Postnikova, Polina and Kudinsky, Daniil and Smirnov, Alexander and Sukhinina, Anastasia and Lila, Alexander and Kurmukov, Anvar},
    title     = {RheumaVIT: Transformer-Based Model for Automated Scoring of Hand Joints in Rheumatoid Arthritis},
    booktitle = {Proceedings of the IEEE/CVF International Conference on Computer Vision (ICCV) Workshops},
    month     = {October},
    year      = {2023},
    pages     = {2522-2531}
}

@ARTICLE{siméoni2025dinov3,
       author = {{Sim{\'e}oni}, Oriane and {Vo}, Huy V. and {Seitzer}, Maximilian and {Baldassarre}, Federico and {Oquab}, Maxime and {Jose}, Cijo and {Khalidov}, Vasil and {Szafraniec}, Marc and {Yi}, Seungeun and {Ramamonjisoa}, Micha{\"e}l and {Massa}, Francisco and {Haziza}, Daniel and {Wehrstedt}, Luca and {Wang}, Jianyuan and {Darcet}, Timoth{\'e}e and {Moutakanni}, Th{\'e}o and {Sentana}, Leonel and {Roberts}, Claire and {Vedaldi}, Andrea and {Tolan}, Jamie and {Brandt}, John and {Couprie}, Camille and {Mairal}, Julien and {J{\'e}gou}, Herv{\'e} and {Labatut}, Patrick and {Bojanowski}, Piotr},
        title = "{DINOv3}",
      journal = {arXiv e-prints},
         year = 2025,
        month = aug,
          eid = {arXiv:2508.10104},
        pages = {arXiv:2508.10104},
          doi = {10.48550/arXiv.2508.10104},
archivePrefix = {arXiv},
       eprint = {2508.10104},
 primaryClass = {cs.CV},
       adsurl = {https://ui.adsabs.harvard.edu/abs/2025arXiv250810104S}
}

@article{anhui,
  author    = {Zijian Wang and Jian Liu and Zongyun Gu and Chuanfu Li},
  title     = {An Efficient CNN for Hand X-Ray Overall Scoring of Rheumatoid Arthritis},
  journal   = {Complexity},
  volume    = {2022},
  number    = {1},
  pages     = {5485606},
  year      = {2022},
  doi       = {10.1155/2022/5485606},
  url       = {https://onlinelibrary.wiley.com/doi/10.1155/2022/5485606},
}

@InProceedings{anhui_baseline2,
  author    = {Zhiyan Bo and Laura C. Coates and Bart{\l}omiej W. Papie{\.{z}}},
  editor    = {Moi Hoon Yap and Connah Kendrick and Ardhendu Behera and Timothy Cootes and Reyer Zwiggelaar},
  title     = {Deep Learning Models to Automate the Scoring of Hand Radiographs for Rheumatoid Arthritis},
  booktitle = {Medical Image Understanding and Analysis},
  year      = {2024},
  publisher = {Springer Nature Switzerland},
  address   = {Cham},
  pages     = {398--413},
  isbn      = {978-3-031-66958-3}
}

@InProceedings{ema,
    author="Gao, Ning
    and Zhou, Sanping
    and Wang, Le
    and Zheng, Nanning",
    editor="Leonardis, Ale{\v{s}}
    and Ricci, Elisa
    and Roth, Stefan
    and Russakovsky, Olga
    and Sattler, Torsten
    and Varol, G{\"u}l",
    title="PMT: Progressive Mean Teacher via Exploring Temporal Consistency for Semi-Supervised Medical Image Segmentation",
    booktitle="Computer Vision -- ECCV 2024",
    year="2025",
    publisher="Springer Nature Switzerland",
    address="Cham",
    pages="144--160",
    isbn="978-3-031-73113-6"
}

@inproceedings{mlr,
  author    = {Yibing Wei and Abhinav Gupta and Pedro Morgado},
  title     = {Towards Latent Masked Image Modeling for Self-Supervised Visual Representation Learning},
  booktitle = {Computer Vision -- ECCV 2024: Proceedings, Part XXXIX},
  year      = {2024},
  pages     = {1--17},
  publisher = {Springer-Verlag},
  address   = {Berlin, Heidelberg},
  doi       = {10.1007/978-3-031-72933-1_1},
  url       = {https://doi.org/10.1007/978-3-031-72933-1_1},
  location  = {Milan, Italy},
  isbn      = {978-3-031-72932-4}
}

@article{tian2025yolov12,
  title={YOLOv12: Attention-Centric Real-Time Object Detectors},
  author={Tian, Yunjie and Ye, Qixiang and Doermann, David},
  journal={arXiv preprint arXiv:2502.12524},
  year={2025}
}

@inproceedings{10.1007/978-3-030-58452-8_13,
author = {Carion, Nicolas and Massa, Francisco and Synnaeve, Gabriel and Usunier, Nicolas and Kirillov, Alexander and Zagoruyko, Sergey},
title = {End-to-End Object Detection with Transformers},
year = {2020},
isbn = {978-3-030-58451-1},
publisher = {Springer-Verlag},
address = {Berlin, Heidelberg},
url = {https://doi.org/10.1007/978-3-030-58452-8_13},
doi = {10.1007/978-3-030-58452-8_13},
booktitle = {Computer Vision – ECCV 2020: 16th European Conference, Glasgow, UK, August 23–28, 2020, Proceedings, Part I},
pages = {213–229},
numpages = {17},
location = {Glasgow, United Kingdom}
}

@ARTICLE{huang2024deim,
       author = {{Huang}, Shihua and {Lu}, Zhichao and {Cun}, Xiaodong and {Yu}, Yongjun and {Zhou}, Xiao and {Shen}, Xi},
        title = "{DEIM: DETR with Improved Matching for Fast Convergence}",
      journal = {arXiv e-prints},
         year = 2024,
        month = dec,
          eid = {arXiv:2412.04234},
        pages = {arXiv:2412.04234},
          doi = {10.48550/arXiv.2412.04234},
archivePrefix = {arXiv},
       eprint = {2412.04234},
 primaryClass = {cs.CV},
       adsurl = {https://ui.adsabs.harvard.edu/abs/2024arXiv241204234H}
}

@ARTICLE{huang2025deimv2,
       author = {{Huang}, Shihua and {Hou}, Yongjie and {Liu}, Longfei and {Yu}, Xuanlong and {Shen}, Xi},
        title = "{Real-Time Object Detection Meets DINOv3}",
      journal = {arXiv e-prints},
         year = 2025,
        month = sep,
          eid = {arXiv:2509.20787},
        pages = {arXiv:2509.20787},
          doi = {10.48550/arXiv.2509.20787},
archivePrefix = {arXiv},
       eprint = {2509.20787},
 primaryClass = {cs.CV},
       adsurl = {https://ui.adsabs.harvard.edu/abs/2025arXiv250920787H}
}

@article{AEbonet,
  author    = {Mojtaba Sirati-Amsheh and Elham Shabaninia and Ali Chaparian},
  title     = {AE-BoNet: A Deep Learning Method for Pediatric Bone Age Estimation Using an Unsupervised Pre-Trained Model},
  journal   = {Journal of Biomedical Physics and Engineering},
  year      = {2025},
  volume    = {15},
  number    = {3},
  pages     = {271--280},
  doi       = {10.31661/jbpe.v0i0.2304-1609},
  pmid      = {40510303},
  pmc       = {PMC12153479},
  publisher = {Isfahan University of Medical Sciences},
  address   = {Iran},
  month     = jun,
  note      = {eCollection 2025 Jun}
}

@article{ccrbaa,
  author    = {Zhangyong Li and Wang Chen and Yang Ju and Yong Chen and Zhengjun Hou and Xinwei Li and Yuhao Jiang},
  title     = {Bone Age Assessment Based on Deep Neural Networks with Annotation-Free Cascaded Critical Bone Region Extraction},
  journal   = {Frontiers in Artificial Intelligence},
  year      = {2023},
  volume    = {6},
  pages     = {1142895},
  doi       = {10.3389/frai.2023.1142895},
  pmid      = {36937708},
  pmc       = {PMC10017763},
  publisher = {Frontiers Media S.A.},
  address   = {Switzerland},
  month     = mar,
  note      = {eCollection 2023}
}

@article{rhpeconvnext,
  author    = {Joshua T. Bram and Ayoosh Pareek and Samuel A. Beber and Ruth H. Jones and M. Moein Shariatnia and Amir Daliliyazdi and Olivia C. Tracey and Daniel W. Green and Peter D. Fabricant},
  title     = {Determination of Skeletal Age From Hand Radiographs Using Deep Learning},
  journal   = {The American Journal of Sports Medicine},
  year      = {2025},
  volume    = {53},
  number    = {11},
  pages     = {2715--2725},
  doi       = {10.1177/03635465251359618},
  pmid      = {40815847},
  pmc       = {PMC12381387},
  publisher = {SAGE Publications},
  address   = {United States},
  month     = sep,
  note      = {Epub 2025 Aug 15}
}

@inproceedings{bonet,
  author    = {Mar{\'i}a Escobar and Cristina Gonz{\'a}lez and Felipe Torres and Laura Daza and Gustavo Triana and Pablo Arbel{\'a}ez},
  title     = {Hand Pose Estimation for Pediatric Bone Age Assessment},
  booktitle = {Medical Image Computing and Computer Assisted Intervention – MICCAI 2019: Proceedings, Part VI},
  year      = {2019},
  pages     = {531--539},
  publisher = {Springer-Verlag},
  address   = {Berlin, Heidelberg},
  doi       = {10.1007/978-3-030-32226-7_59},
  url       = {https://doi.org/10.1007/978-3-030-32226-7_59},
  isbn      = {978-3-030-32225-0},
  location  = {Shenzhen, China}
}

@inproceedings{st_ssad,
  author    = {Antoine Spahr and Behzad Bozorgtabar and Jean-Philippe Thiran},
  title     = {Self-Taught Semi-Supervised Anomaly Detection on Upper Limb X-Rays},
  booktitle = {Proceedings of the 2021 IEEE 18th International Symposium on Biomedical Imaging (ISBI)},
  year      = {2021},
  pages     = {1632--1636},
  doi       = {10.1109/ISBI48211.2021.9433771},
  publisher = {IEEE}
}

@article{mura_d169,
  author    = {Anna Solovyova and Igor Solovyov},
  title     = {X-Ray Bone Abnormalities Detection Using MURA Dataset},
  journal   = {arXiv preprint arXiv:2008.03356},
  year      = {2020},
  url       = {https://arxiv.org/abs/2008.03356}
}

@article{dha,
  author    = {Arkadiusz Gertych and Aifeng Zhang and James Sayre and Sylwia Pospiech-Kurkowska and H. K. Huang},
  title     = {Bone Age Assessment of Children Using a Digital Hand Atlas},
  journal   = {Computerized Medical Imaging and Graphics},
  year      = {2007},
  volume    = {31},
  number    = {4--5},
  pages     = {322--331},
  doi       = {10.1016/j.compmedimag.2007.02.012},
  pmid      = {17387000},
  pmc       = {PMC1978493},
  publisher = {Elsevier},
  address   = {United States},
  month     = jun
}

@misc{hf_vit_cxr,
  author       = {{chanelcolgate}},
  title        = {vit-base-patch16-224-chest-x-ray: fine-tuned ViT-Base on a chest-x-ray classification dataset},
  howpublished = {\url{https://huggingface.co/chanelcolgate/vit-base-patch16-224-chest-x-ray}},
  year         = {2025},
}

@misc{medpix,
  author       = {{National Library of Medicine}},
  title        = {{MedPix{\textregistered}: Database of Medical Images, Teaching Cases, and Clinical Topics}},
  howpublished = {\url{https://medpix.nlm.nih.gov/}},
  note         = {Accessed: 2026-03-06}
}

@book{terminologia_anatomica,
  author    = {{Federative International Programme for Anatomical Terminology (FIPAT)}},
  title     = {Terminologia Anatomica: International Anatomical Terminology},
  edition   = {2nd},
  year      = {2019},
  publisher = {Thieme Medical Publishers},
  address   = {Stuttgart, Germany}
}

@article{radlex,
  author    = {Curtis P. Langlotz},
  title     = {RadLex: A New Method for Indexing Online Educational Materials},
  journal   = {Radiographics},
  year      = {2006},
  volume    = {26},
  number    = {6},
  pages     = {1595--1597},
  doi       = {10.1148/rg.266065168},
  pmid      = {17102038},
  publisher = {Radiological Society of North America},
  address   = {United States},
  month     = nov
}

@article{moradmand2025,
  author    = {Hajar Moradmand and Lei Ren},
  title     = {Multistage Deep Learning Methods for Automating Radiographic Sharp Score Prediction in Rheumatoid Arthritis},
  journal   = {Scientific Reports},
  year      = {2025},
  volume    = {15},
  number    = {1},
  pages     = {3391},
  doi       = {10.1038/s41598-025-86073-0},
  url       = {https://doi.org/10.1038/s41598-025-86073-0},
  publisher = {Nature Publishing Group},
  address   = {United Kingdom},
  month     = jan
}

@article{Siragusa_medpix_2025,
   title={MedPix 2.0: A Comprehensive Multimodal Biomedical Data Set for Advanced AI Applications with Retrieval Augmented Generation and Knowledge Graphs},
   ISSN={2364-1541},
   url={http://dx.doi.org/10.1007/s41019-025-00297-8},
   DOI={10.1007/s41019-025-00297-8},
   journal={Data Science and Engineering},
   publisher={Springer Science and Business Media LLC},
   author={Siragusa, Irene and Contino, Salvatore and Ciura, Massimo La and Alicata, Rosario and Pirrone, Roberto},
   year={2025},
   month=jul }

@article{radclip,
  title        = {{RadCLIP}: Enhancing Radiologic Image Analysis Through Contrastive Language-Image Pretraining},
  author       = {Lu, Zhixiu and Li, Hailong and Parikh, Nehal A. and Dillman, Jonathan R. and He, Lili},
  journal      = {IEEE Transactions on Neural Networks and Learning Systems},
  year         = {2025},
  month        = oct,
  volume       = {36},
  number       = {10},
  pages        = {17613--17622},
  doi          = {10.1109/TNNLS.2025.3568036},
  pmid         = {40434863},
  pmcid        = {PMC12498476}
}

@article{dosovitskiy2020vit,
  title={An image is worth 16x16 words: Transformers for image recognition at scale},
  author={Dosovitskiy, Alexey and Beyer, Lucas and Kolesnikov, Alexander and Weissenborn, Dirk and Zhai, Xiaohua and Unterthiner, Thomas and Dehghani, Mostafa and Minderer, Matthias and Heigold, Georg and Gelly, Sylvain and others},
  journal={arXiv preprint arXiv:2010.11929},
  year={2020}
}

@inproceedings{touvron2021deit,
  title={Training data-efficient image transformers \& distillation through attention},
  author={Touvron, Hugo and Cord, Matthieu and Douze, Matthijs and Massa, Francisco and Sablayrolles, Alexandre and J{\'e}gou, Herv{\'e}},
  booktitle={International conference on machine learning},
  pages={10347--10357},
  year={2021},
  organization={PMLR}
}

@article{kim2026generalizable,
  title={A generalizable large-scale foundation model for musculoskeletal radiographs},
  author={Kim, Shinn and Lee, Soobin and Shin, Kyoungseob and Kim, Han-Soo and Kim, Yongsung and Kim, Minsu and Nam, Juhong and Ko, Somang and Kwon, Daeheon and Huh, Wook and others},
  journal={arXiv preprint arXiv:2602.03076},
  year={2026}
}

@article{singla2025multi,
  title={Multi Anatomy X-Ray Foundation Model},
  author={Singla, Nishank and Koos, Krisztian and Haddadpour, Farzin and Shandiz, Amin Honarmandi and Chum, Lovish and Xu, Xiaojian and Jin, Qing and Bas, Erhan},
  journal={arXiv preprint arXiv:2509.12146},
  year={2025}
}

@inproceedings{tarvainen2017meanteacher,
  title={Mean Teachers Are Better Role Models:
         Weight-Averaged Consistency Targets
         Improve Semi-Supervised Deep Learning Results},
  author={Tarvainen, Antti and Valpola, Harri},
  booktitle={Advances in Neural Information Processing Systems},
  volume={30},
  year={2017}
}

@inproceedings{grill2020byol,
  title={Bootstrap Your Own Latent: A New Approach
         to Self-Supervised Learning},
  author={Grill, Jean-Bastien and Strub, Florian
          and Altch{\'e}, Florent and Tallec, Corentin
          and Richemond, Pierre H. and Buchatskaya, Elena
          and Doersch, Carl and Pires, Bernardo Avila
          and Guo, Zhaohan Daniel and Azar, Mohammad Gheshlaghi
          and Piot, Bilal and Kavukcuoglu, Koray
          and Munos, R{\'e}mi and Valko, Michal},
  booktitle={Advances in Neural Information Processing Systems},
  volume={33},
  year={2020}
}

@inproceedings{caron2021dino,
  title={Emerging Properties in Self-Supervised Vision Transformers},
  author={Caron, Mathilde and Touvron, Hugo
          and Misra, Ishan and J{\'e}gou, Herv{\'e}
          and Mairal, Julien and Bojanowski, Piotr
          and Joulin, Armand},
  booktitle={Proceedings of the IEEE/CVF International
             Conference on Computer Vision},
  pages={9650--9660},
  year={2021}
}
\bibliographystyle{plainnat}





\clearpage
\appendix

\begin{center}
{\Large \textbf{Supplementary Material}}

\end{center}
\renewcommand{\thefigure}{\thesection\arabic{figure}}
\renewcommand{\thetable}{\thesection\arabic{table}}

\renewcommand{\thesubsection}{\thesection.\arabic{subsection}}

\begin{center}
\Large \textbf{Table of Contents}
\end{center}

{
\setlength{\parskip}{0.35em}   
\setlength{\parindent}{0pt}    
\renewcommand{\baselinestretch}{1.08}\selectfont  

\textbf{A. Pretrain Configuration and Hyperparameters} \dotfill \pageref{supp:pretrain_config} \par
\hspace*{2em}A.1.\ Student and Teacher Backbones \dotfill \pageref{supp:stu_bone} \par
\hspace*{3em}Student encoder \dotfill \pageref{supp:stu_encoder} \par
\hspace*{3em}Expert teachers \dotfill \pageref{supp:expert} \par
\hspace*{2em}A.2.\ Pretraining Data and Augmentation \dotfill \pageref{supp:augmentation} \par
\hspace*{3em}Data sources \dotfill \pageref{supp:data_src} \par
\hspace*{3em}Image preprocessing \dotfill \pageref{supp:preprocess} \par
\hspace*{2em}A.3.\ Masking and Latent Regressor \dotfill \pageref{supp:mlr} \par
\hspace*{3em}Patch masking \dotfill \pageref{supp:patch_msk} \par
\hspace*{3em}Masked token regressor \dotfill \pageref{supp:regressor} \par
\hspace*{2em}A.4.\ Losses, EMA, and Optimization \dotfill \pageref{supp:loss_ema_opt} \par
\hspace*{3em}Latent alignment and reconstruction \dotfill \pageref{supp:align_recon} \par
\hspace*{3em}EMA teacher \dotfill \pageref{supp:ema} \par
\hspace*{2em}A.5.\ Ablation Pretraining Variants \dotfill \pageref{supp:variant} \par

\vspace{0.4em}
\textbf{B. Results Details} \dotfill \pageref{supp:resuls_details} \par
\hspace*{2em}B.1.\ Joint Detection Results \dotfill \pageref{supp:joint_det} \par
\hspace*{2em}B.2.\ Bone Segmentation Results \dotfill \pageref{supp:bone_seg} \par
\hspace*{2em}B.3.\ Ablation on Expert Composition \dotfill \pageref{supp:abla_expert} \par
\hspace*{2em}B.4.\ Ablation on Loss Composition \dotfill \pageref{supp:abla_loss} \par
\hspace*{2em}B.5.\ Ablation on Masking Strategy \dotfill \pageref{supp:abla_mask} \par
\hspace*{2em}B.6.\ Effect of Higher-Resolution Downstream Fine-Tuning \dotfill 
\pageref{supp:appendix_resolution_ablation} \par
\hspace*{2em}B.7.\ Full Fine-Tuning Protocol and Frozen-Backbone Analysis \dotfill \pageref{supp:appendix_frozen_backbone} \par
\hspace*{2em}B.8.\ Multi-Seed Stability Analysis \dotfill \pageref{supp:appendix_multiseed_airinii}  \par
\hspace*{2em}B.9.\ VQA Case Analysis \dotfill \pageref{supp:vqa} 
\par

\vspace{0.4em}
\textbf{C. Dataset Details} \dotfill \pageref{supp:dataset_details} \par
\hspace*{2em}C.1.\ Public Datasets \dotfill \pageref{supp:public} \par
\hspace*{2em}C.2.\ Newly Annotated Datasets \dotfill \pageref{supp:new} \par

\vspace{0.4em}
\textbf{D. Limitations and Future Work} \dotfill \pageref{supp:limitations_future} \par

\vspace{0.4em}
\textbf{E. Reproducibility Check} \dotfill \pageref{supp:reproducibility} \par
\hspace*{2em}E.1.\ Compute \dotfill \pageref{supp:repro_compute} \par
\hspace*{2em}E.2.\ Data organization and preprocessing \dotfill \pageref{supp:repro_data} \par
\hspace*{2em}E.3.\ Training configuration \dotfill \pageref{supp:repro_train} \par
\hspace*{2em}E.4.\ Code entry points and artifacts \dotfill \pageref{supp:repro_code} \par

\vspace{0.4em}
\textbf{F. Broader Impact} \dotfill \pageref{supp:broader} \par

}

\par\vspace{2em}\hrule
\clearpage


\section{Pretrain Configuration and Hyperparameters}
\label{supp:pretrain_config}

This section records the implementation settings needed to reproduce HandXFM pretraining. We focus on the components and configuration of the proposed method; the learning objectives themselves are introduced in~\Cref{sec:method}.

\subsection{Student and Teacher Backbones}
\label{supp:stu_bone}

\paragraph{Student encoder.}\label{supp:stu_encoder}
HandXFM initializes its student from a DINOv3 ViT-B/16 checkpoint using the Hugging Face \texttt{AutoModel} interface. The original ViT architecture is retained: $16\times16$ patches, 12 transformer layers, 12 attention heads, and a hidden width of 768 (MLP ratio 4). At the pretraining resolution of $224\times224$, the model produces 196 patch tokens and one \texttt{[CLS]} token. When the checkpoint contains register tokens, these are excluded from the tokens used in the distillation objectives. The resulting encoder has approximately 86M parameters. The final exponential-moving-average (EMA) encoder, rather than the instantaneous student, supplies the default weights for downstream experiments.

The $224\times224$ pretraining resolution limits the computational cost of multi-expert supervision; downstream tasks may use higher input resolutions as specified in their respective experimental protocols. The student architecture is not modified: HandXFM learns a hand-radiograph representation through its pretraining objectives.

\paragraph{Expert teachers.}\label{supp:expert}
Two pretrained visual encoders supply complementary latent supervision. The BiomedCLIP ViT-B/16 image encoder is loaded with \texttt{open\_clip.create\_model\_and\_transforms} and its associated preprocessing. Its visual representations were learned through biomedical image--text contrastive pretraining, providing language-aligned visual features without requiring text inputs during HandXFM pretraining. A chest-radiograph ViT-B/16, loaded with \texttt{AutoModel} and its \texttt{AutoImageProcessor}, contributes a distinct radiography-domain appearance prior.

Here, \emph{frozen} means that neither expert receives gradient updates during HandXFM pretraining. Both operate only as feature-target generators; the student, the expert-specific linear projection layers $P_j$, and the masked latent regressor are optimized. Each $P_j$ maps the corresponding expert representation to the 768-dimensional student space and is instantiated as a \texttt{LazyLinear} layer. Teacher targets are organized into a global \texttt{[CLS]} representation and a spatial token sequence. For patch-wise supervision, a teacher sequence with a different number of spatial positions is truncated or padded to the student's 196-token grid.

\noindent\textbf{Radiography-domain teacher provenance.}
The radiography-domain teacher uses the following Hugging Face checkpoint:
\par\smallskip\noindent{\small\texttt{chanelcolgate/vit-base-patch16-224-chest-x-ray}}\par\smallskip
Starting from a standard ViT-B/16 checkpoint, this model was fine-tuned for binary normal-versus-pneumonia classification on the public trpakov chest X-ray classification collection. According to its model card, this source dataset has 12,230 training, 1,165 validation, and 582 test images. This teacher was not trained to identify hand joints or bone boundaries. Its role in HandXFM is to provide features adapted to projection radiographs, whereas BiomedCLIP supplies representations shaped by biomedical image--text learning. We do not assume an exclusive semantic/structural division between the teachers; the corresponding single- and dual-teacher comparisons are reported in~\Cref{sec:ablation}.

\subsection{Pretraining Data and Augmentation}
\label{supp:augmentation}

\paragraph{Data sources.}\label{supp:data_src}
We curated approximately 30K hand radiographs from
six publicly available datasets: RSNA Bone Age,
RHPE, MURA \texttt{XR\_HAND}, AIRINIIReumHands,
Anhui RA, and MedPix Hand. HandXFM uses images
from the designated training partitions for
unlabeled pretraining, with validation and test
images excluded. Radiographs are processed as
whole-hand inputs rather than cropped into
individual joint or bone regions. The dataset
partitions used for downstream training and
evaluation are detailed in
\Cref{tab:dataset_allocation}.

\paragraph{Image preprocessing.} \label{supp:preprocess}
The notation $x\in\mathbb{R}^{3\times224\times224}$ in~\Cref{sec:method} denotes the standardized network input, not three independently acquired X-ray channels. Images from Anhui RA and MedPix may be encoded as RGB; grayscale radiographs from RSNA, RHPE, MURA, and AIRINIIReumHands are represented by repeating their intensity channel across the three RGB channels. This permits use of the pretrained three-channel ViT encoders and does not introduce additional color information.

Augmentations are applied before feature extraction. Following a resize to $256\times256$, each input undergoes a randomly applied horizontal flip ($p=0.5$) and a rotation bounded by $\pm10^\circ$ with zero padding; the central $224\times224$ region is retained afterward. \texttt{ToTensor} converts pixel intensities to $[0,1]$. Backbone-specific input normalization is subsequently handled by the relevant model processor; for the DINOv3 processor, \texttt{do\_rescale=False} avoids rescaling tensors a second time. BiomedCLIP uses the preprocessing associated with its \texttt{open\_clip} configuration.

\subsection{Masking and Latent Regressor} \label{supp:mlr}

\paragraph{Patch masking.} \label{supp:patch_msk}
HandXFM masks 50\% of the $14\times14$ patch locations. For each image, we aggregate grayscale Sobel-gradient magnitudes into 196 patch-level values via adaptive average pooling. The resulting grid biases the masking distribution toward edge-rich locations; it is combined with uniform sampling according to~\Cref{sec:method}. In the reported HandXFM configuration, \texttt{mask\_type=edge}, \texttt{mask\_ratio=0.5}, \texttt{hw=14}, and \texttt{mask\_edge\_alpha=0.5}. The coefficient $\alpha=0.5$ is the weight of the uniform component; patches with stronger edges are more likely to be \emph{masked} and predicted. Patches are selected without replacement.

\paragraph{Masked token regressor.} \label{supp:regressor}
The latent regressor $\mathcal{R}_{\phi}$ uses learned mask queries to attend to visible student patch features and estimate representations at masked locations. Its architecture has six transformer blocks, hidden width 768, eight attention heads, MLP ratio 4, GELU activations, LayerNorm, and linearly increasing stochastic depth up to 0.1. Patch locations are specified using fixed 2D sine--cosine embeddings of size $196\times768$, assigned to both the context features and the mask queries. The predictions are trained against expert-projected latent targets, not reconstructed image pixels.

\subsection{Losses, EMA, and Optimization} \label{supp:loss_ema_opt}

\paragraph{Latent alignment and reconstruction.} \label{supp:align_recon}
The visible-token alignment and masked-token reconstruction terms use mean-squared differences between student representations (or regressor predictions) and the corresponding projected expert targets. The contributions from the available experts are averaged, and an additional \texttt{[CLS]} alignment term supervises the global representation. In the objective defined in~\Cref{sec:method}, the main-run coefficients are $\lambda_{\mathrm{align}}=1$, $\lambda_{\mathrm{recon}}=1$, $\lambda_{\mathrm{ema}}=0.01$, and $\lambda_{\mathrm{cls}}=0.01$.

The loss coefficients were selected by examining
the relative magnitudes of the individual terms
during preliminary pretraining. Alignment and
reconstruction had comparable scales and were
assigned unit weights, whereas the auxiliary EMA
and CLS terms were assigned smaller coefficients
to prevent them from dominating the objective.
These weights were kept fixed for pretraining
runs using the complete objective.

\paragraph{EMA teacher.} \label{supp:ema}
An EMA copy of the student is maintained with momentum $\mu=0.996$ and updated after each optimization step as $\theta_m\leftarrow\mu\theta_m+(1-\mu)\theta$. Its parameters are never updated by backpropagation. The EMA features supply a temporal consistency target for the current student, with the corresponding contribution controlled by $\lambda_{\mathrm{ema}}=0.01$. Weight-averaged teachers are established techniques in self-training and self-distillation, including Mean Teacher~\cite{tarvainen2017meanteacher}, BYOL~\cite{grill2020byol}, and DINO~\cite{caron2021dino}; EMA is used here for stabilization and is not claimed as an independently novel algorithmic component.

\paragraph{Optimizer and learning rates.} \label{supp:opt}
We use \texttt{AdamW} to optimize the student encoder and the regressor/projection modules in separate parameter groups, while the frozen experts and EMA encoder are excluded from gradient-based updates. The encoder and head learning rates are $3\times10^{-5}$ and $1\times10^{-4}$, respectively. The AdamW coefficients are $(\beta_1,\beta_2)=(0.9,0.999)$, with weight decay set to zero. The complete main-run settings are listed in~\Cref{tab:handxfm_pretrain_config}.

\paragraph{Training schedule.} \label{supp:schedule}
Pretraining runs for 10 epochs with batch size 512. The student encoder is frozen during the first two epochs, then unfrozen; the trainable projection and regression modules are optimized throughout. The final EMA encoder is exported for downstream adaptation. For reproducibility, the saved training state records both encoder copies, the latent-prediction modules, optimizer variables, epoch numbers, and aggregate losses.

\begin{table}[t]
\centering
\caption{HandXFM pretraining configuration. All values describe the main setting; ablations modify only the indicated component.}
\label{tab:handxfm_pretrain_config}
\normalsize
\setlength{\tabcolsep}{4pt}
\renewcommand{\arraystretch}{1.40}
\begin{tabular}{@{}p{0.42\linewidth}p{0.54\linewidth}@{}}
\toprule
Component & Configuration \\
\midrule
Student & DINOv3 ViT-B/16; 12 blocks; 12 heads; $D=768$ \\
Input resolution / grid & $224\times224$ / $14\times14$ ($N=196$) \\
Masking & Edge-aware; ratio $0.5$; $\alpha=0.5$ \\
Latent regressor & 6 blocks; $D=768$; 8 heads; MLP ratio 4 \\
Regressor regularization & GELU; LayerNorm; drop path up to 0.1 \\
Loss weights & $(\lambda_{\mathrm{align}},\lambda_{\mathrm{recon}},\lambda_{\mathrm{ema}},\lambda_{\mathrm{cls}})=(1,1,0.01,0.01)$ \\
EMA momentum & $0.996$ \\
Optimizer & \texttt{AdamW} \\
Learning rates (student / heads) & $3\times10^{-5}$ / $1\times10^{-4}$ \\
Weight decay & $0$ \\
AdamW betas $(\beta_1,\beta_2)$ & $(0.9,0.999)$ \\
Epochs / batch size & 10 / 512 \\
Student warmup freeze & 2 epochs \\
Workers / logging / checkpointing & 8 / 100 iterations / 5 epochs \\
\bottomrule
\end{tabular}
\end{table}

\subsection{Ablation Pretraining Variants}
\label{supp:variant}

The teacher-composition, objective, and masking
ablations use the same student initialization,
pretraining data, optimization settings, and
downstream evaluation protocols unless otherwise
specified. The single-teacher configurations
receive latent supervision from either BiomedCLIP
or the chest-radiograph encoder, while the
dual-teacher configuration uses both. The
objective ablations isolate visible-token
alignment and masked latent reconstruction.
For the masking ablation, uniformly sampled
patches replace edge-biased masking while
maintaining the same masking ratio of 0.5.


\section{Results Details}
\label{supp:resuls_details}
\subsection{Joint detection result}\label{supp:joint_det}
\begin{figure*}[t]
\centering
\newlength{\BThreeNaturalHeight}
\settoheight{\BThreeNaturalHeight}{\includegraphics{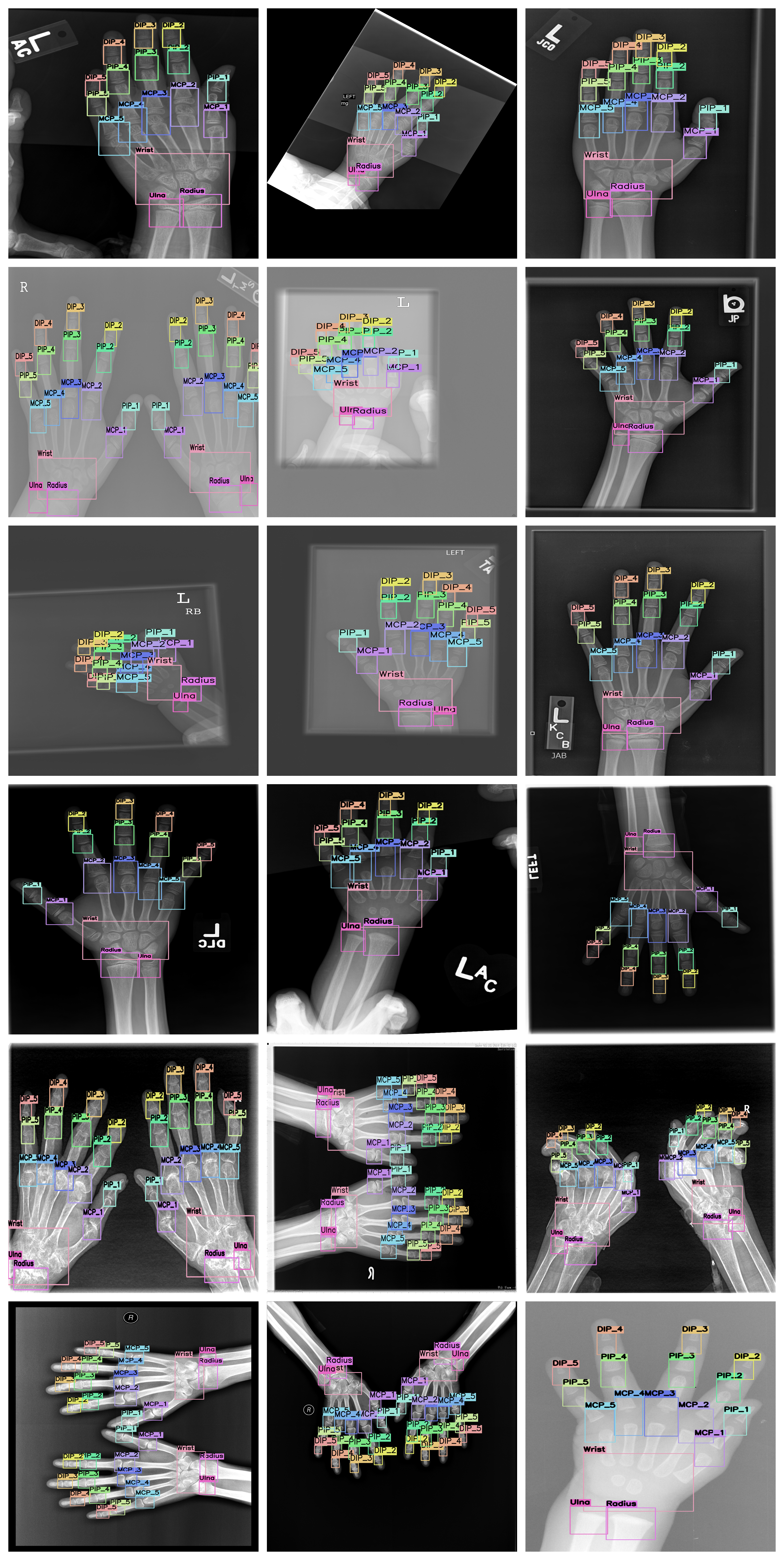}}
\includegraphics[width=\textwidth,trim=0pt {\dimexpr\BThreeNaturalHeight/3\relax} 0pt 0pt,clip,keepaspectratio]{figs/joint_det_result_grid.png}
\caption{Representative predictions on the held-out RSNA-based joint-detection evaluation subset. The panels illustrate localization under changes in hand orientation, finger configuration, and radiographic appearance.}
\label{fig:joint_det_result_grid}
\end{figure*}

\Cref{fig:joint_det_result_grid} illustrates predictions for the 17 anatomical targets across representative hand radiographs. The examples include variation in positioning, finger overlap, and local anatomy. The detector operates on full radiographs without requiring a separate image-registration step. These visual examples supplement the quantitative localization metrics in the main paper; they are not, on their own, a measurement of robustness under every acquisition condition.

\subsection{Bone segmentation result} \label{supp:bone_seg}

\begin{figure*}[t]
\centering
\newlength{\BFourNaturalHeight}
\settoheight{\BFourNaturalHeight}{\includegraphics{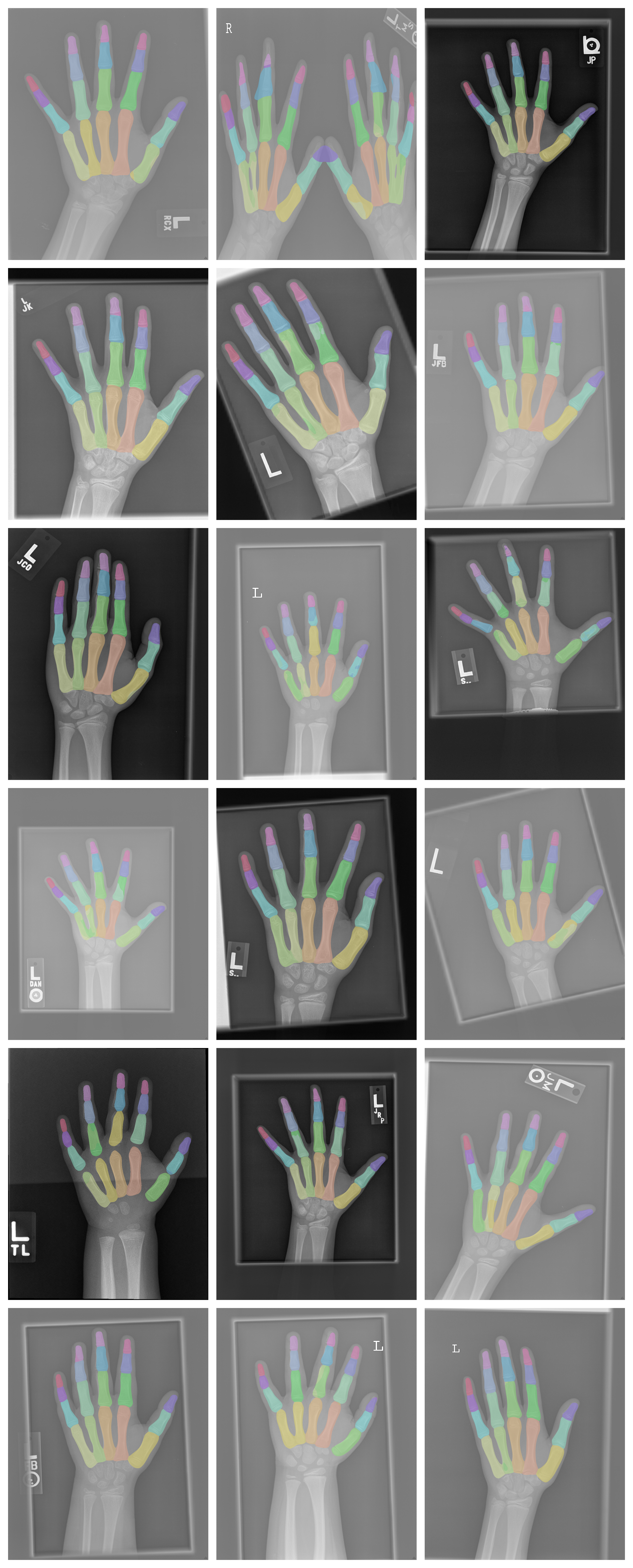}}
\includegraphics[width=\textwidth,trim=0pt {\dimexpr\BFourNaturalHeight/2\relax} 0pt 0pt,clip,keepaspectratio]{figs/bone_seg_grid.png}
\caption{Qualitative 19-class bone-segmentation results on held-out radiographs. The color-coded masks distinguish individual annotated bone categories and illustrate predictions across varying image appearances and hand configurations.}
\label{fig:bone_seg_grid}
\end{figure*}

The examples in \Cref{fig:bone_seg_grid} visualize the correspondence between the radiographs and their predicted anatomical masks. These results provide examples of instance-aware bone delineation across subjects; the quantitative instance-localization and pixel-level comparisons are reported separately. The proposed method is intended to identify individual bone categories, rather than merely segmenting the aggregate bone foreground.

\begin{table}[t]
\centering
\caption{Pixel-level bone-segmentation comparison with conventional medical semantic-segmentation baselines on the same evaluation set. Results complement the instance-level COCO-style evaluation; pixel accuracy alone can be influenced by background prevalence. Higher is better for all metrics.}
\label{tab:supp_semantic_seg_baselines}
\normalsize
\setlength{\tabcolsep}{4pt}
\renewcommand{\arraystretch}{1.4}
\begin{tabular*}{\linewidth}{@{\extracolsep{\fill}}lcccc@{}}
\toprule
Method & Pixel ACC & Precision & Recall & F1 \\
\midrule
nnU-Net & 0.9587 & 0.4384 & 0.4762 & 0.4538 \\
U-Net & \textbf{0.9597} & 0.4549 & 0.4803 & 0.4645 \\
HandXFM & 0.9586 & \textbf{0.4596} & \textbf{0.4984} & \textbf{0.4777} \\
\bottomrule
\end{tabular*}
\end{table}

To complement the instance-level baselines in the main paper, \Cref{tab:supp_semantic_seg_baselines} compares HandXFM with U-Net and nnU-Net using pixel-wise measures. HandXFM obtains the highest precision, recall, and F1 in this comparison, whereas U-Net achieves slightly higher pixel accuracy. These results address a different aspect of performance from the reported COCO-style instance mAP; the pixel metrics do not by themselves evaluate the correct separation of neighboring bone instances.

\subsection{Ablation on Expert Composition} \label{supp:abla_expert}
\paragraph{Effect of expert on regression tasks.}
\label{supp:abla_expert_reg}
\Cref{tab:supp_ablation_expert_reg} compares the DINOv3-only control with BiomedCLIP-only, chest-radiograph-only, and dual-teacher supervision under the same downstream protocols. The effects depend on the target: The chest-only variant is stronger than BiomedCLIP-only on the RSNA and RHPE bone-age metrics and on several AIRINIIReumHands measures. The dual-teacher model leads on RSNA, RHPE, Anhui, and erosion metrics, as well as AIRINIIReumHands JSN PCC; for JSN RMSE, MAE, and $R^2$, chest-only performs better. Thus, the teacher contributions are complementary but cannot be assigned exclusively semantic or anatomical roles.

\begin{table}[h]
\centering
\caption{Ablation on expert composition---regression. Lower is better for RMSE/MAE; higher is better for $R^2$/PCC.}
\label{tab:supp_ablation_expert_reg}
\small
\setlength{\tabcolsep}{3pt}
\renewcommand{\arraystretch}{1.25}

\begin{tabular*}{\linewidth}{@{\extracolsep{\fill}}llcccc@{}}
\toprule
Dataset & Metric & DINOv3 & BiomedCLIP & Chest X-ray & HandXFM \\
\midrule
\multicolumn{6}{c}{\textbf{Bone Age}} \\
\midrule
RSNA & RMSE $\downarrow$ & 12.702 & 10.497 & 10.178 & \textbf{9.932} \\
     & MAE $\downarrow$  & 9.106 & 7.683 & 7.584 & \textbf{7.343} \\
     & $R^2$ $\uparrow$ & 0.907 & 0.937 & 0.940 & \textbf{0.943} \\
     & PCC $\uparrow$   & 0.953 & 0.968 & 0.970 & \textbf{0.972} \\
\addlinespace[2pt]
RHPE & RMSE $\downarrow$ & 27.828 & 18.487 & 12.462 & \textbf{11.073} \\
     & MAE $\downarrow$  & 21.856 & 13.484 & 9.388 & \textbf{8.363} \\
     & $R^2$ $\uparrow$ & 0.550 & 0.801 & 0.910 & \textbf{0.929} \\
     & PCC $\uparrow$   & 0.790 & 0.911 & 0.956 & \textbf{0.966} \\
\midrule
\multicolumn{6}{c}{\textbf{SvH Scoring}} \\
\midrule
AIRINII (Erosion) & RMSE $\downarrow$ & 0.525 & 0.420 & 0.306 & \textbf{0.298} \\
                 & MAE $\downarrow$  & 0.300 & 0.230 & 0.144 & \textbf{0.111} \\
                 & $R^2$ $\uparrow$ & 0.127 & 0.440 & 0.704 & \textbf{0.719} \\
                 & PCC $\uparrow$   & 0.426 & 0.758 & 0.894 & \textbf{0.929} \\
\addlinespace[2pt]
AIRINII (JSN) & RMSE $\downarrow$ & 0.714 & 0.738 & \textbf{0.591} & 0.705 \\
              & MAE $\downarrow$  & 0.558 & 0.575 & \textbf{0.380} & 0.608 \\
              & $R^2$ $\uparrow$ & 0.591 & 0.562 & \textbf{0.720} & 0.601 \\
              & PCC $\uparrow$   & 0.803 & 0.782 & 0.870 & \textbf{0.885} \\
\addlinespace[2pt]
Anhui (SvH) & RMSE $\downarrow$ & 31.232 & 26.115 & 20.567 & \textbf{18.231} \\
            & MAE $\downarrow$  & 20.301 & 17.094 & 13.529 & \textbf{12.356} \\
            & $R^2$ $\uparrow$ & 0.447 & 0.613 & 0.760 & \textbf{0.812} \\
            & PCC $\uparrow$   & 0.702 & 0.796 & 0.884 & \textbf{0.903} \\
\bottomrule
\end{tabular*}
\end{table}

\paragraph{Effect of expert on  classification task.} \label{supp:abla_expert_cls}
On MURA (\Cref{tab:supp_ablation_expert_mura}), the dual-teacher model records the highest AUC, accuracy, and precision. The chest-only variant has the highest F1, while DINOv3-only has the highest recall. These differences show a trade-off between precision and sensitivity; an overall improvement in AUC should not be interpreted as superiority on every operating-point metric.

\begin{table*}[t]
\centering
\caption{Ablation on expert composition---MURA normal/abnormal classification. Higher is better for all metrics.}
\label{tab:supp_ablation_expert_mura}
\scriptsize
\setlength{\tabcolsep}{4pt}
\renewcommand{\arraystretch}{1.08}
\resizebox{\linewidth}{!}{%
\begin{tabular}{llcccc}
\toprule
& Metric & DINOv3 Only & BiomedCLIP Only & Chest X-ray Only & Dual (HandXFM) \\
\midrule
& AUC $\uparrow$ & 0.7405 & 0.7771 & 0.8485 & \textbf{0.8599} \\
& ACC $\uparrow$ & 0.6741 & 0.6987 & 0.7768 & \textbf{0.7857} \\
& F1 $\uparrow$ & 0.6589 & 0.6650 & \textbf{0.6933} & 0.6903 \\
& P $\uparrow$ & 0.5900 & 0.6262 & 0.8248 & \textbf{0.8843} \\
& R $\uparrow$ & \textbf{0.7460} & 0.7090 & 0.5979 & 0.5661 \\
\bottomrule
\end{tabular}%
}
\end{table*}

\paragraph{Effect of expert on joint detection.} \label{supp:abla_expert_det}
The matched joint-localization comparison in \Cref{tab:supp_ablation_expert_joint} is particularly informative: BiomedCLIP-only falls below the DINOv3-only baseline on all three reported metrics, whereas chest-only yields modest gains on the aggregate and stricter-IoU metrics. Dual-teacher HandXFM substantially improves mAP@0.5:0.95 and mAP@0.75. These results establish that the benefit arises from the combination of teachers, rather than assuming that every individual teacher improves every task.

\begin{table*}[t]
\centering
\caption{Ablation on expert composition---17-class joint detection. Higher is better for all metrics.}
\label{tab:supp_ablation_expert_joint}
\normalsize
\setlength{\tabcolsep}{8pt}
\renewcommand{\arraystretch}{1.6}

\begin{tabular*}{\textwidth}{@{\extracolsep{\fill}}lcccc@{}}
\toprule
Metric & DINOv3 Only & BiomedCLIP Only & Chest X-ray Only & Dual (Ours) \\
\midrule
mAP@0.5:0.95 $\uparrow$ & 0.5900 & 0.5550 & 0.6017 & \textbf{0.7636} \\
mAP@0.5 $\uparrow$ & 0.9650 & 0.9515 & 0.9621 & \textbf{0.9802} \\
mAP@0.75 $\uparrow$ & 0.6487 & 0.5810 & 0.6771 & \textbf{0.9026} \\
\bottomrule
\end{tabular*}
\end{table*}

\paragraph{Effect of expert on bone segmentation.}\label{supp:abla_expert_seg}
The segmentation ablation (\Cref{tab:supp_ablation_expert_bone}) separates instance classification, instance localization, and pixel-level segmentation. Dual-teacher supervision leads on instance classification and on the two stricter localization summaries (mAP@0.5:0.95 and mAP@0.75). However, chest-only has the highest mAP@0.5, and BiomedCLIP-only attains higher pixel accuracy and recall than the dual model. This pattern supports the value of combined supervision for bone-level localization without implying a uniform advantage on all pixel metrics.

\begin{table*}[t]
\centering
\caption{Ablation on expert composition---bone segmentation, including instance-level classification, instance localization, and pixel-level metrics. Higher is better.
The DINOv3-only configuration is the
no-expert-teacher pretraining ablation,
rather than the native DINOv3 checkpoint
evaluated in the main baseline comparison.}
\label{tab:supp_ablation_expert_bone}

\normalsize
\setlength{\tabcolsep}{6pt}
\renewcommand{\arraystretch}{1.40}

\begin{tabular*}{\textwidth}{@{\extracolsep{\fill}}lcccc@{}}
\toprule
Metric & DINOv3 Only & BiomedCLIP Only & Chest X-ray Only & Dual (HandXFM) \\
\midrule

\multicolumn{5}{c}{\textbf{Instance-level Classification}} \\
\midrule
ACC $\uparrow$ & 0.9474 & 0.9732 & 0.9675 & \textbf{0.9751} \\
P $\uparrow$   & 0.9435 & 0.9445 & 0.9428 & \textbf{0.9481} \\
R $\uparrow$   & 0.9000 & 0.9245 & 0.9191 & \textbf{0.9264} \\
F1 $\uparrow$  & 0.9210 & 0.9344 & 0.9307 & \textbf{0.9371} \\

\midrule
\multicolumn{5}{c}{\textbf{Instance-level Localization}} \\
\midrule
mAP@0.5:0.95 $\uparrow$ & 0.3121 & 0.4800 & 0.5187 & \textbf{0.6101} \\
mAP@0.5 $\uparrow$      & 0.6998 & 0.9165 & \textbf{0.9632} & 0.9466 \\
mAP@0.75 $\uparrow$     & 0.2092 & 0.4397 & 0.5349 & \textbf{0.7030} \\

\midrule
\multicolumn{5}{c}{\textbf{Pixel-level Segmentation}} \\
\midrule
ACC $\uparrow$ & 0.9512 & \textbf{0.9593} & 0.9589 & 0.9586 \\
P $\uparrow$   & 0.4174 & 0.4582 & 0.4528 & \textbf{0.4596} \\
R $\uparrow$   & 0.4850 & \textbf{0.5026} & 0.5024 & 0.4984 \\
F1 $\uparrow$  & 0.4436 & \textbf{0.4777} & 0.4753 & \textbf{0.4777} \\
\bottomrule
\end{tabular*}
\end{table*}

\subsection{Ablation on Loss Composition } \label{supp:abla_loss}
\paragraph{Effect of loss on regression tasks.}
\label{supp:abla_loss_reg}
The regression results in \Cref{tab:supp_ablation_loss_reg} indicate that the preferred objective depends on the benchmark and metric. The combined formulation performs best on RSNA, RHPE, and Anhui. For AIRINIIReumHands erosion, alignment-only has the lowest RMSE and MAE and the highest $R^2$, whereas the combined objective has the highest PCC. For JSN, reconstruction-only has the lowest errors while the combined formulation again has the highest PCC. This is evidence for cross-task complementarity, not a claim that the combined loss dominates every measurement.

\begin{table*}[t]
\centering
\caption{Ablation on loss composition---regression. Lower is better for RMSE/MAE; higher is better for $R^2$/PCC.}
\label{tab:supp_ablation_loss_reg}

\normalsize
\setlength{\tabcolsep}{5pt}
\renewcommand{\arraystretch}{1.35}

\begin{tabular*}{\textwidth}{@{\extracolsep{\fill}}llccc@{}}
\toprule
Dataset & Metric & Alignment Only & Reconstruction Only & Combined \\
\midrule

\multicolumn{5}{c}{\textbf{Bone Age Prediction}} \\
\midrule
RSNA & RMSE $\downarrow$ & 10.200 & 10.415 & \textbf{9.932} \\
     & MAE $\downarrow$  & 7.600 & 7.676 & \textbf{7.343} \\
     & $R^2$ $\uparrow$ & 0.940 & 0.938 & \textbf{0.943} \\
     & PCC $\uparrow$   & 0.970 & 0.968 & \textbf{0.972} \\

\addlinespace[3pt]
RHPE & RMSE $\downarrow$ & 11.622 & 12.329 & \textbf{11.073} \\
     & MAE $\downarrow$  & 8.907 & 9.372 & \textbf{8.363} \\
     & $R^2$ $\uparrow$ & 0.921 & 0.912 & \textbf{0.929} \\
     & PCC $\uparrow$   & 0.964 & 0.957 & \textbf{0.966} \\

\midrule
\multicolumn{5}{c}{\textbf{SvH Scoring}} \\
\midrule
AIRINII (Erosion) & RMSE $\downarrow$ & \textbf{0.223} & 0.240 & 0.298 \\
                 & MAE $\downarrow$  & \textbf{0.104} & 0.141 & 0.111 \\
                 & $R^2$ $\uparrow$ & \textbf{0.843} & 0.818 & 0.719 \\
                 & PCC $\uparrow$   & 0.919 & 0.906 & \textbf{0.929} \\

\addlinespace[3pt]
AIRINII (JSN) & RMSE $\downarrow$ & 0.804 & \textbf{0.649} & 0.705 \\
              & MAE $\downarrow$  & 0.601 & \textbf{0.416} & 0.608 \\
              & $R^2$ $\uparrow$ & 0.481 & \textbf{0.662} & 0.601 \\
              & PCC $\uparrow$   & 0.777 & 0.859 & \textbf{0.885} \\

\addlinespace[3pt]
Anhui (SvH) & RMSE $\downarrow$ & 20.987 & 21.194 & \textbf{18.231} \\
            & MAE $\downarrow$  & 13.453 & 13.939 & \textbf{12.356} \\
            & $R^2$ $\uparrow$ & 0.750 & 0.745 & \textbf{0.812} \\
            & PCC $\uparrow$   & 0.887 & 0.880 & \textbf{0.903} \\

\bottomrule
\end{tabular*}
\end{table*}

\paragraph{Effect of loss on classification task.} \label{supp:abla_loss_cls}
In MURA classification (\Cref{tab:supp_ablation_loss_mura}), the joint objective produces the highest AUC, accuracy, and precision, whereas reconstruction-only yields the highest F1 and recall. The distinction matters because threshold-dependent metrics can respond differently to the same learned representation.

\begin{table*}[t]
\centering
\caption{Ablation on loss composition---MURA normal/abnormal classification. Higher is better for all metrics.}
\label{tab:supp_ablation_loss_mura}

\normalsize
\setlength{\tabcolsep}{5pt}
\renewcommand{\arraystretch}{1.35}

\begin{tabular*}{\textwidth}{@{\extracolsep{\fill}}lccc@{}}
\toprule
Metric & Alignment Only & Reconstruction Only & Combined \\
\midrule
AUC $\uparrow$ & 0.8413 & 0.8430 & \textbf{0.8599} \\
ACC $\uparrow$ & 0.7790 & 0.7723 & \textbf{0.7857} \\
F1 $\uparrow$  & 0.6937 & \textbf{0.7243} & 0.6903 \\
P $\uparrow$   & 0.8629 & 0.7403 & \textbf{0.8843} \\
R $\uparrow$   & 0.5661 & \textbf{0.7090} & 0.5661 \\
\bottomrule
\end{tabular*}
\end{table*}

\paragraph{Effect of loss on joint detection.} \label{supp:abla_loss_det}
For detection, reconstruction-only exceeds alignment-only on each reported mAP measure, while the combined loss provides a larger improvement at the stricter IoU threshold (\Cref{tab:supp_ablation_loss_joint}). This suggests that the two terms interact constructively for localization, but the experiment does not isolate the individual contribution of the auxiliary EMA or CLS terms.

\begin{table*}[t]
\centering
\caption{Ablation on loss composition---17-class joint detection. Higher is better for all metrics.}
\label{tab:supp_ablation_loss_joint}

\normalsize
\setlength{\tabcolsep}{5pt}
\renewcommand{\arraystretch}{1.35}

\begin{tabular*}{\textwidth}{@{\extracolsep{\fill}}lccc@{}}
\toprule
Metric & Alignment Only & Reconstruction Only & Combined \\
\midrule
mAP@0.5:0.95 $\uparrow$ & 0.5765 & 0.5995 & \textbf{0.7636} \\
mAP@0.5 $\uparrow$      & 0.9624 & 0.9702 & \textbf{0.9802} \\
mAP@0.75 $\uparrow$     & 0.6230 & 0.6699 & \textbf{0.9026} \\
\bottomrule
\end{tabular*}
\end{table*}

\paragraph{Effect of loss on bone segmentation.} \label{supp:abla_loss_seg}
The bone-segmentation comparison in \Cref{tab:supp_ablation_loss_bone} is likewise metric dependent. The combined objective leads for instance-level accuracy, recall, F1, and all reported instance-localization measures. Alignment-only provides higher pixel accuracy, precision, recall, and F1 than the combined model, while reconstruction-only ties the combined model on instance precision. Accordingly, the main benefit of combining these objectives is observed in instance discrimination and localization rather than in every pixel-wise score.

\begin{table}[t]
\centering
\caption{Ablation on loss composition---bone segmentation, including instance and pixel-level metrics. Higher is better.}
\label{tab:supp_ablation_loss_bone}

\fontsize{10.5pt}{13pt}\selectfont
\setlength{\tabcolsep}{4pt}
\renewcommand{\arraystretch}{1.45}

\begin{tabular*}{\linewidth}{@{\extracolsep{\fill}}lccc@{}}
\toprule
Metric & Alignment Only & Reconstruction Only & Combined \\
\midrule

\multicolumn{4}{c}{\textbf{Instance-level Classification}} \\
\midrule
ACC $\uparrow$ & 0.9722 & 0.9512 & \textbf{0.9751} \\
P $\uparrow$   & 0.9436 & \textbf{0.9481} & \textbf{0.9481} \\
R $\uparrow$   & 0.9236 & 0.9036 & \textbf{0.9264} \\
F1 $\uparrow$  & 0.9334 & 0.9251 & \textbf{0.9371} \\

\midrule
\multicolumn{4}{c}{\textbf{Instance-level Localization}} \\
\midrule
mAP@0.5:0.95 $\uparrow$ & 0.5071 & 0.2444 & \textbf{0.6101} \\
mAP@0.5 $\uparrow$      & 0.9107 & 0.6872 & \textbf{0.9466} \\
mAP@0.75 $\uparrow$     & 0.5056 & 0.0984 & \textbf{0.7030} \\

\midrule
\multicolumn{4}{c}{\textbf{Pixel-level Segmentation}} \\
\midrule
ACC $\uparrow$ & \textbf{0.9602} & 0.9600 & 0.9586 \\
P $\uparrow$   & \textbf{0.4628} & 0.4616 & 0.4596 \\
R $\uparrow$   & \textbf{0.5016} & 0.4974 & 0.4984 \\
F1 $\uparrow$  & \textbf{0.4804} & 0.4769 & 0.4777 \\
\bottomrule
\end{tabular*}
\end{table}

\subsection{Ablation on Masking Strategy} \label{supp:abla_mask}
\paragraph{Effect of masking strategy on regression tasks.}
As shown in \Cref{tab:supp_ablation_mask_reg}, edge-aware masking improves the main RSNA, RHPE, and Anhui bone-age or SvH regression measures. On the smaller AIRINIIReumHands dataset, the trend is mixed: random masking has slightly higher erosion PCC and lower JSN RMSE/MAE, whereas edge-aware masking leads on erosion RMSE/$R^2$ and JSN PCC. The edge-biased sampling policy therefore offers a generally useful inductive bias, with exceptions that motivate the multi-seed analysis in \Cref{supp:appendix_multiseed_airinii}.

\begin{table*}[t]
\centering
\caption{Ablation on masking strategy---regression. Lower is better for RMSE/MAE; higher is better for $R^2$/PCC.}
\label{tab:supp_ablation_mask_reg}

\normalsize
\setlength{\tabcolsep}{5pt}
\renewcommand{\arraystretch}{1.35}

\begin{tabular*}{\textwidth}{@{\extracolsep{\fill}}llcc@{}}
\toprule
Dataset & Metric & Random Masking & Edge-aware Masking \\
\midrule

\multicolumn{4}{c}{\textbf{Bone Age Prediction}} \\
\midrule
RSNA & RMSE $\downarrow$ & 10.729 & \textbf{9.932} \\
     & MAE $\downarrow$  & 7.870 & \textbf{7.343} \\
     & $R^2$ $\uparrow$ & 0.934 & \textbf{0.943} \\
     & PCC $\uparrow$   & 0.967 & \textbf{0.972} \\

\addlinespace[3pt]
RHPE & RMSE $\downarrow$ & 11.650 & \textbf{11.073} \\
     & MAE $\downarrow$  & 8.789 & \textbf{8.363} \\
     & $R^2$ $\uparrow$ & 0.921 & \textbf{0.929} \\
     & PCC $\uparrow$   & 0.962 & \textbf{0.966} \\

\midrule
\multicolumn{4}{c}{\textbf{SvH Scoring}} \\
\midrule
AIRINII (Erosion) & RMSE $\downarrow$ & 0.438 & \textbf{0.298} \\
                  & MAE $\downarrow$  & 0.127 & \textbf{0.111} \\
                  & $R^2$ $\uparrow$ & 0.391 & \textbf{0.719} \\
                  & PCC $\uparrow$   & \textbf{0.941} & 0.929 \\

\addlinespace[3pt]
AIRINII (JSN) & RMSE $\downarrow$ & \textbf{0.689} & 0.705 \\
              & MAE $\downarrow$  & \textbf{0.550} & 0.608 \\
              & $R^2$ $\uparrow$ & \textbf{0.618} & 0.601 \\
              & PCC $\uparrow$   & 0.812 & \textbf{0.885} \\

\addlinespace[3pt]
Anhui (SvH) & RMSE $\downarrow$ & 21.402 & \textbf{18.231} \\
            & MAE $\downarrow$  & 13.911 & \textbf{12.356} \\
            & $R^2$ $\uparrow$ & 0.740 & \textbf{0.812} \\
            & PCC $\uparrow$   & 0.884 & \textbf{0.903} \\

\bottomrule
\end{tabular*}
\end{table*}

\paragraph{Effect of masking strategy on classification tasks.}
On MURA, edge-aware masking improves all five reported classification measures, including AUC (0.7739 to 0.8599) and F1 (0.5860 to 0.6903), relative to random masking (\Cref{tab:supp_ablation_mask_mura}). This result is consistent with a benefit from emphasizing radiographically informative boundaries during pretraining, although the evaluation does not identify which specific image regions cause the improvement.

\begin{table}[t]
\centering
\caption{Ablation on masking strategy---MURA normal/abnormal classification. Higher is better for all metrics.}
\label{tab:supp_ablation_mask_mura}

\normalsize
\setlength{\tabcolsep}{5pt}
\renewcommand{\arraystretch}{1.50}

\begin{tabular*}{\linewidth}{@{\extracolsep{\fill}}lcc@{}}
\toprule
Metric & Random Masking & Edge-aware Masking \\
\midrule
AUC $\uparrow$ & 0.7739 & \textbf{0.8599} \\
ACC $\uparrow$ & 0.7098 & \textbf{0.7857} \\
F1 $\uparrow$  & 0.5860 & \textbf{0.6903} \\
P $\uparrow$   & 0.7360 & \textbf{0.8843} \\
R $\uparrow$   & 0.4868 & \textbf{0.5661} \\
\bottomrule
\end{tabular*}
\end{table}

\paragraph{Effect of masking strategy on joint detection.}
For joint detection (\Cref{tab:supp_ablation_mask_joint}), the greatest improvement appears under stricter localization criteria: mAP@0.75 rises from 0.8485 to 0.9026 and average mAP@0.5:0.95 from 0.7125 to 0.7636. The random-mask model is marginally higher at mAP@0.5 (0.9844 versus 0.9802), indicating that the advantage of edge-aware masking is most visible for tighter overlap requirements.

\begin{table}[t]
\centering
\caption{Ablation on masking strategy---17-class joint detection. Higher is better for all metrics.}
\label{tab:supp_ablation_mask_joint}

\normalsize
\setlength{\tabcolsep}{5pt}
\renewcommand{\arraystretch}{1.60}

\begin{tabular*}{\linewidth}{@{\extracolsep{\fill}}lcc@{}}
\toprule
Metric & Random Masking & Edge-aware Masking \\
\midrule
mAP@0.5:0.95 $\uparrow$ & 0.7125 & \textbf{0.7636} \\
mAP@0.5 $\uparrow$      & \textbf{0.9844} & 0.9802 \\
mAP@0.75 $\uparrow$     & 0.8485 & \textbf{0.9026} \\
\bottomrule
\end{tabular*}
\end{table}

\paragraph{Effect of masking strategy on bone segmentation.}
For the segmentation task, edge-aware masking improves all reported instance classification and localization scores (\Cref{tab:supp_ablation_mask_bone}); the mAP@0.75 increase is especially pronounced (0.4617 to 0.7030). Pixel-level results show a different balance: the random-mask variant has higher pixel accuracy, recall, and F1, whereas edge-aware masking has higher pixel precision. These measurements support the localization-oriented role of the proposed masking policy without implying that it optimizes every foreground-pixel metric.

\begin{table}[t]
\centering
\caption{Ablation on masking strategy---bone segmentation, including instance and pixel-level measures. Higher is better.}
\label{tab:supp_ablation_mask_bone}

\normalsize
\setlength{\tabcolsep}{4pt}
\renewcommand{\arraystretch}{1.40}

\begin{tabular*}{\linewidth}{@{\extracolsep{\fill}}lcc@{}}
\toprule
Metric & Random Masking & Edge-aware Masking \\
\midrule

\multicolumn{3}{c}{\textbf{Instance-level Classification}} \\
\midrule
ACC $\uparrow$ & 0.9713 & \textbf{0.9751} \\
P $\uparrow$   & 0.9419 & \textbf{0.9481} \\
R $\uparrow$   & 0.9227 & \textbf{0.9264} \\
F1 $\uparrow$  & 0.9321 & \textbf{0.9371} \\

\midrule
\multicolumn{3}{c}{\textbf{Instance-level Localization}} \\
\midrule
mAP@0.5:0.95 $\uparrow$ & 0.4680 & \textbf{0.6101} \\
mAP@0.5 $\uparrow$      & 0.8914 & \textbf{0.9466} \\
mAP@0.75 $\uparrow$     & 0.4617 & \textbf{0.7030} \\

\midrule
\multicolumn{3}{c}{\textbf{Pixel-level Segmentation}} \\
\midrule
ACC $\uparrow$ & \textbf{0.9596} & 0.9586 \\
P $\uparrow$   & 0.4555 & \textbf{0.4596} \\
R $\uparrow$   & \textbf{0.5178} & 0.4984 \\
F1 $\uparrow$  & \textbf{0.4836} & 0.4777 \\
\bottomrule
\end{tabular*}
\end{table}

\subsection{Effect of Higher-Resolution Downstream Fine-Tuning} \label{supp:appendix_resolution_ablation}
HandXFM is pretrained with $224\times224$ inputs. Downstream tasks can use a larger input grid by interpolating the pretrained positional embeddings with two-dimensional bicubic interpolation, a standard ViT adaptation practice \cite{dosovitskiy2020vit,touvron2021deit}. This strategy permits evaluation of resolution-dependent features without changing the pretraining corpus or retraining the foundation model.

In the Anhui whole-hand SvH experiment, only the downstream fine-tuning resolution is changed (\Cref{tab:anhui_resolution_ablation}). Performance improves from $224\times224$ to $384\times384$, reducing RMSE from 18.231 to 15.195 and increasing PCC from 0.903 to 0.933. At $448\times448$, performance remains higher than the $224\times224$ configuration but is lower than the $384\times384$ result. We therefore report a task-specific benefit of moderately higher downstream resolution, not a general claim that increasing resolution always improves performance.

\begin{table}[t]
\centering
\caption{Effect of downstream input resolution on Anhui whole-hand SvH scoring. All variants share the $224\times224$ pretrained backbone and differ only in fine-tuning resolution.}
\label{tab:anhui_resolution_ablation}

\normalsize
\setlength{\tabcolsep}{4pt}
\renewcommand{\arraystretch}{1.55}

\begin{tabular*}{\linewidth}{@{\extracolsep{\fill}}lcccc@{}}
\toprule
Input Resolution & RMSE $\downarrow$ & MAE $\downarrow$ & $R^2 \uparrow$ & PCC $\uparrow$ \\
\midrule
$224\times224$ & 18.231 & 12.356 & 0.812 & 0.903 \\
$384\times384$ & \textbf{15.195} & \textbf{10.513} & \textbf{0.869} & \textbf{0.933} \\
$448\times448$ & 16.404 & 11.601 & 0.847 & 0.928 \\
\bottomrule
\end{tabular*}
\end{table}

\subsection{Full Fine-Tuning Protocol and Frozen-Backbone Analysis}
\label{supp:appendix_frozen_backbone}
Unless explicitly noted, the principal foundation-model comparisons use end-to-end fine-tuning with each method initialized from its own released checkpoint and trained on the same downstream data partition and task-specific evaluation protocol. Thus, the main results reflect adapted models, not frozen-backbone linear probes. The exception analyzed here holds the visual encoder fixed while fitting the same lightweight regression head on Anhui whole-hand SvH scoring.

As reported in \Cref{tab:anhui_frozen_backbone}, HandXFM achieves the highest frozen-backbone PCC (0.684) and $R^2$ (0.333), together with the lowest RMSE and MAE of the evaluated frozen encoders. This provides complementary evidence that the pretrained representation makes task-relevant features accessible without encoder updates. The frozen HandXFM result is nevertheless substantially weaker than its fully fine-tuned performance in the main paper; consequently, the probe should not be interpreted as a replacement for downstream adaptation.

\begin{table}[t]
\centering
\caption{Frozen-backbone comparison on Anhui whole-hand SvH scoring; only the regression head is fitted.}
\label{tab:anhui_frozen_backbone}

\normalsize
\setlength{\tabcolsep}{4pt}
\renewcommand{\arraystretch}{1.50}

\begin{tabular*}{\linewidth}{@{\extracolsep{\fill}}lcccc@{}}
\toprule
Method & RMSE $\downarrow$ & MAE $\downarrow$ & $R^2 \uparrow$ & PCC $\uparrow$ \\
\midrule
ResNet-50 & 35.288 & 23.869 & 0.290 & 0.602 \\
RadFM & 36.717 & 26.092 & 0.236 & 0.521 \\
RadCLIP & 36.579 & 22.830 & 0.241 & 0.654 \\
HandXFM & \textbf{34.284} & \textbf{21.534} & \textbf{0.333} & \textbf{0.684} \\
\bottomrule
\end{tabular*}
\end{table}

\subsection{Multi-Seed Stability on AIRINIIReumHands}
\label{supp:appendix_multiseed_airinii}
Because AIRINIIReumHands contains relatively few labeled radiographs, single-run ablations can be sensitive to the downstream random seed. We therefore repeated the fixed-split evaluation using five seeds $\{0,1,2,42,3407\}$ and report the mean and standard deviation for each metric in \Cref{tab:airinii_multiseed}. These experiments quantify variability over the selected random seeds; they do not establish uncertainty with respect to patient resampling or changes in clinical site.

HandXFM has the lowest mean erosion RMSE (0.3201) and highest erosion PCC (0.9568). For JSN, it also has the best mean RMSE (0.7067), $R^2$ (0.5810), and PCC (0.8844), although the random-masking variant has a lower mean MAE (0.4692 versus 0.5470). Residual standard deviations are non-negligible---for example, HandXFM JSN PCC varies by $0.0216$ across seeds---so the complete design is best characterized as having stronger mean performance on most reported measures, rather than being uniformly superior or insensitive to initialization.

\begin{table*}[t]
\centering
\caption{Multi-seed evaluation on AIRINIIReumHands over five runs with a fixed split and seeds $\{0,1,2,42,3407\}$. Mean $\pm$ standard deviation.}
\label{tab:airinii_multiseed}

\footnotesize
\setlength{\tabcolsep}{2pt}
\renewcommand{\arraystretch}{1.45}

\resizebox{\textwidth}{!}{%
\begin{tabular}{@{}llcccc@{}}
\toprule
Task & Method & RMSE $\downarrow$ & MAE $\downarrow$ & $R^2 \uparrow$ & PCC $\uparrow$ \\
\midrule

Erosion & RheumaVIT
& $0.4166\pm0.0426$ & $0.2998\pm0.0537$
& $0.4443\pm0.1109$ & $0.7487\pm0.0355$ \\

& RadFM
& $0.3790\pm0.0148$ & $0.2148\pm0.0204$
& $0.5441\pm0.0355$ & $0.7572\pm0.0187$ \\

& RadCLIP
& $0.3729\pm0.0985$ & $0.1308\pm0.0299$
& $0.5287\pm0.2282$ & $0.8969\pm0.0755$ \\

& \shortstack[l]{HandXFM\\(random masking)}
& $0.3256\pm0.1064$ & $0.1248\pm0.0176$
& $0.6281\pm0.2114$ & $0.9207\pm0.0439$ \\

& HandXFM (ours)
& $\mathbf{0.3201\pm0.0507}$ & $\mathbf{0.1236\pm0.0139}$
& $\mathbf{0.6672\pm0.0914}$ & $\mathbf{0.9568\pm0.0137}$ \\

\midrule

JSN & RheumaVIT
& $0.7584\pm0.0563$ & $0.5156\pm0.0484$
& $0.5357\pm0.0718$ & $0.8237\pm0.0244$ \\

& RadFM
& $0.7642\pm0.0306$ & $0.5760\pm0.0180$
& $0.5304\pm0.0378$ & $0.7793\pm0.0223$ \\

& RadCLIP
& $0.8904\pm0.0783$ & $0.7240\pm0.0735$
& $0.3585\pm0.1104$ & $0.7852\pm0.0865$ \\

& \shortstack[l]{HandXFM\\(random masking)}
& $0.7156\pm0.1287$ & $\mathbf{0.4692\pm0.1463}$
& $0.5756\pm0.1635$ & $0.8167\pm0.0276$ \\

& HandXFM (ours)
& $\mathbf{0.7067\pm0.1497}$ & $0.5470\pm0.1375$
& $\mathbf{0.5810\pm0.1738}$ & $\mathbf{0.8844\pm0.0216}$ \\

\bottomrule
\end{tabular}%
}
\end{table*}

\subsection{VQA case analysis}
\label{supp:vqa}

The MedPix-derived closed-set VQA task serves as an exploratory assessment of cross-modal transfer rather than a large-scale clinical VQA benchmark. HandXFM pretraining does not use paired image--text supervision directly. Instead, the student learns from BiomedCLIP's frozen, language-aligned visual encoder. During downstream adaptation, HandXFM image features interact with a frozen BiomedCLIP text encoder through cross-attention, followed by an MLP classifier over a fixed answer vocabulary.

\begin{figure*}[t]
\centering
\footnotesize
\setlength{\tabcolsep}{4pt}
\renewcommand{\arraystretch}{1.04}
\begin{tabular}{@{}p{0.16\textwidth}p{0.20\textwidth}p{0.57\textwidth}@{}}
\toprule
\textbf{Example} &
\textbf{Radiograph} &
\textbf{Question, ground truth, and top-5 predictions} \\
\midrule

\textbf{A. Correct prediction} &
\begin{minipage}[t]{\linewidth}\vspace{0pt}
\centering
\includegraphics[
    width=\linewidth,
    height=0.17\textheight,
    keepaspectratio
]{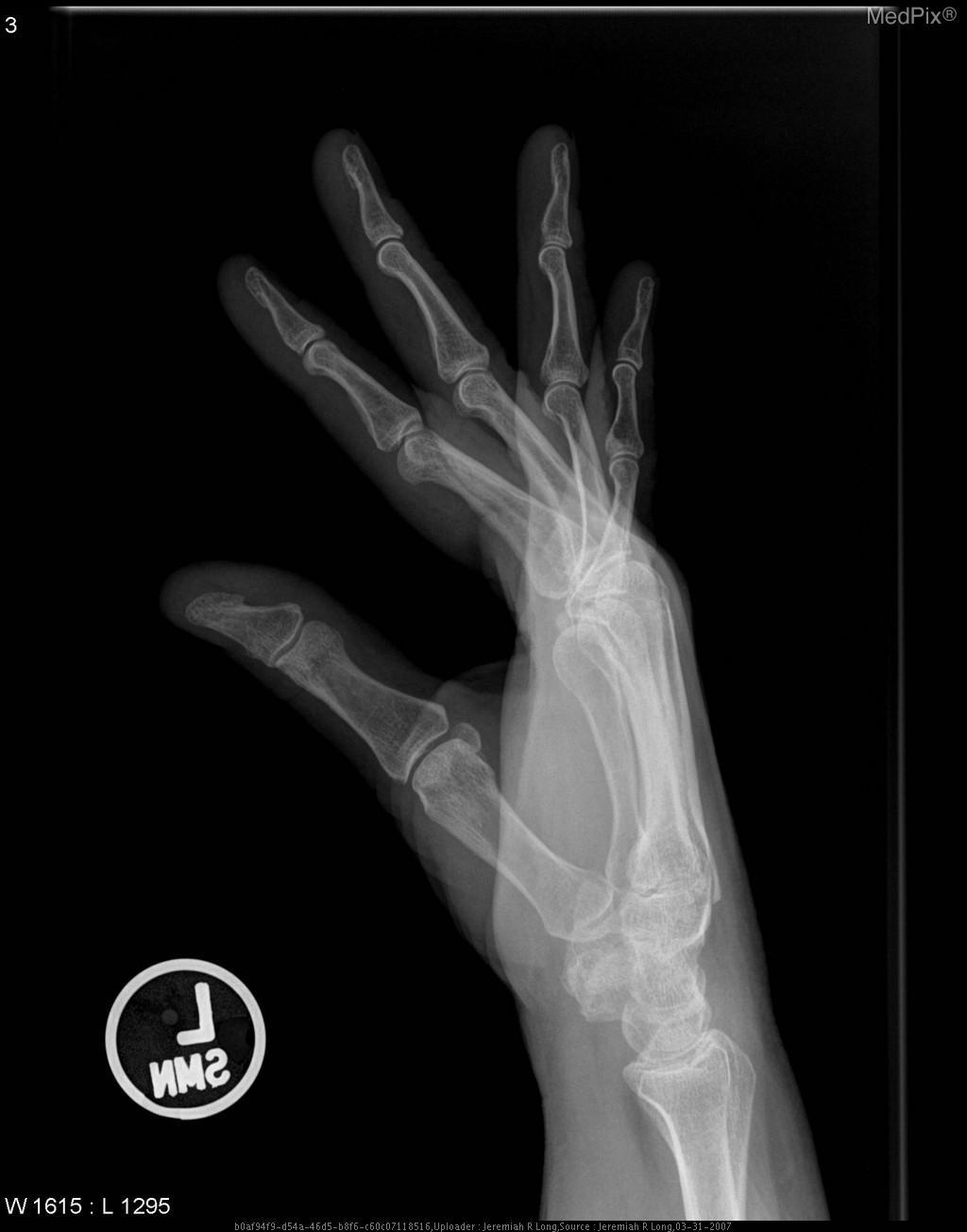}
\end{minipage} &
\begin{minipage}[t]{\linewidth}\vspace{0pt}
\textbf{Question:}
Frontal, lateral and oblique radiographs of the left
hand demonstrate chondrocalcinosis within the
triangular fibrocartilage complex as well as
degenerative changes at the scaphotrapezial joint
evidenced by joint space narrowing and subchondral
sclerosis. The remainder of the osseous and soft
tissue structures are unremarkable.\\[0.3em]
\textbf{GT:} Chondrocalcinosis\\[0.3em]
\textbf{Top-5 predictions:}\\
1.\ Chondrocalcinosis (78.3\%) $\checkmark$\\
2.\ Transscaphoid--transtriquetrial perilunate
fracture dislocation (9.2\%)\\
3.\ Dislocation at the interphalangeal joint
of the thumb with avulsion fracture (3.8\%)\\
4.\ Perilunate dislocation (3.7\%)\\
5.\ Rheumatoid arthritis (0.8\%)
\end{minipage} \\
\midrule

\textbf{B. Correct prediction} &
\begin{minipage}[t]{\linewidth}\vspace{0pt}
\centering
\includegraphics[
    width=\linewidth,
    height=0.17\textheight,
    keepaspectratio
]{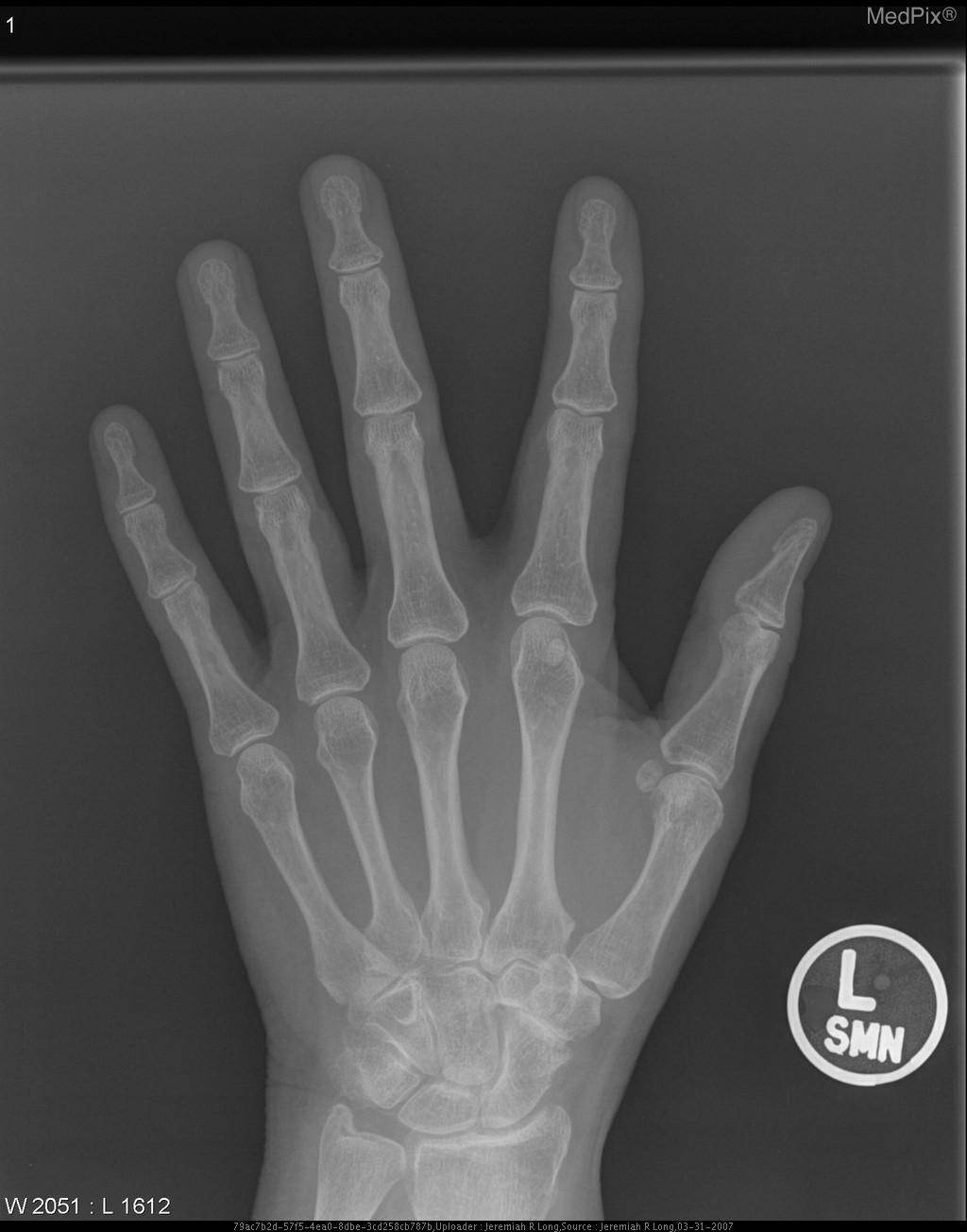}
\end{minipage} &
\begin{minipage}[t]{\linewidth}\vspace{0pt}
\textbf{Question:} Same caption-derived question
as Example A.\\[0.3em]
\textbf{GT:} Chondrocalcinosis\\[0.3em]
\textbf{Top-5 predictions:}\\
1.\ Chondrocalcinosis (84.0\%) $\checkmark$\\
2.\ Transscaphoid--transtriquetrial perilunate
fracture dislocation (7.8\%)\\
3.\ Bilateral Salter Harris Fracture Type 2A
(3.3\%)\\
4.\ Dislocation at the interphalangeal joint
of the thumb with avulsion fracture (0.9\%)\\
5.\ Rheumatoid Arthritis (0.7\%)
\end{minipage} \\
\midrule

\textbf{C. Correct prediction} &
\begin{minipage}[t]{\linewidth}\vspace{0pt}
\centering
\includegraphics[
    width=\linewidth,
    height=0.17\textheight,
    keepaspectratio
]{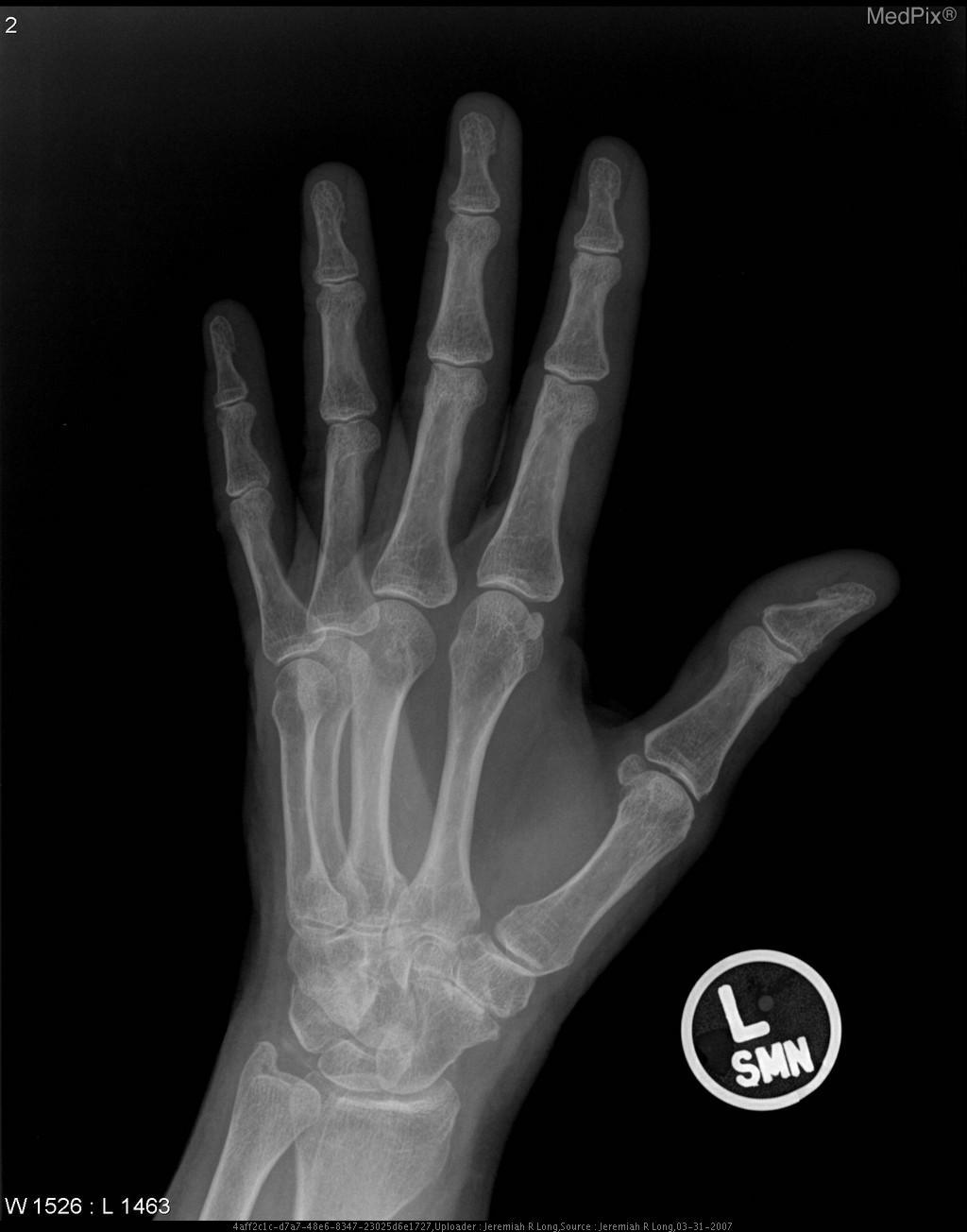}
\end{minipage} &
\begin{minipage}[t]{\linewidth}\vspace{0pt}
\textbf{Question:} Same caption-derived question
as Example A.\\[0.3em]
\textbf{GT:} Chondrocalcinosis\\[0.3em]
\textbf{Top-5 predictions:}\\
1.\ Chondrocalcinosis (84.7\%) $\checkmark$\\
2.\ Transscaphoid--transtriquetrial perilunate
fracture dislocation (9.2\%)\\
3.\ Dislocation at the interphalangeal joint
of the thumb with avulsion fracture (1.4\%)\\
4.\ Bilateral Salter Harris Fracture Type 2A
(1.3\%)\\
5.\ Perilunate dislocation (1.2\%)
\end{minipage} \\
\midrule

\textbf{D. Lower-confidence correct prediction} &
\begin{minipage}[t]{\linewidth}\vspace{0pt}
\centering
\includegraphics[
    width=\linewidth,
    height=0.17\textheight,
    keepaspectratio
]{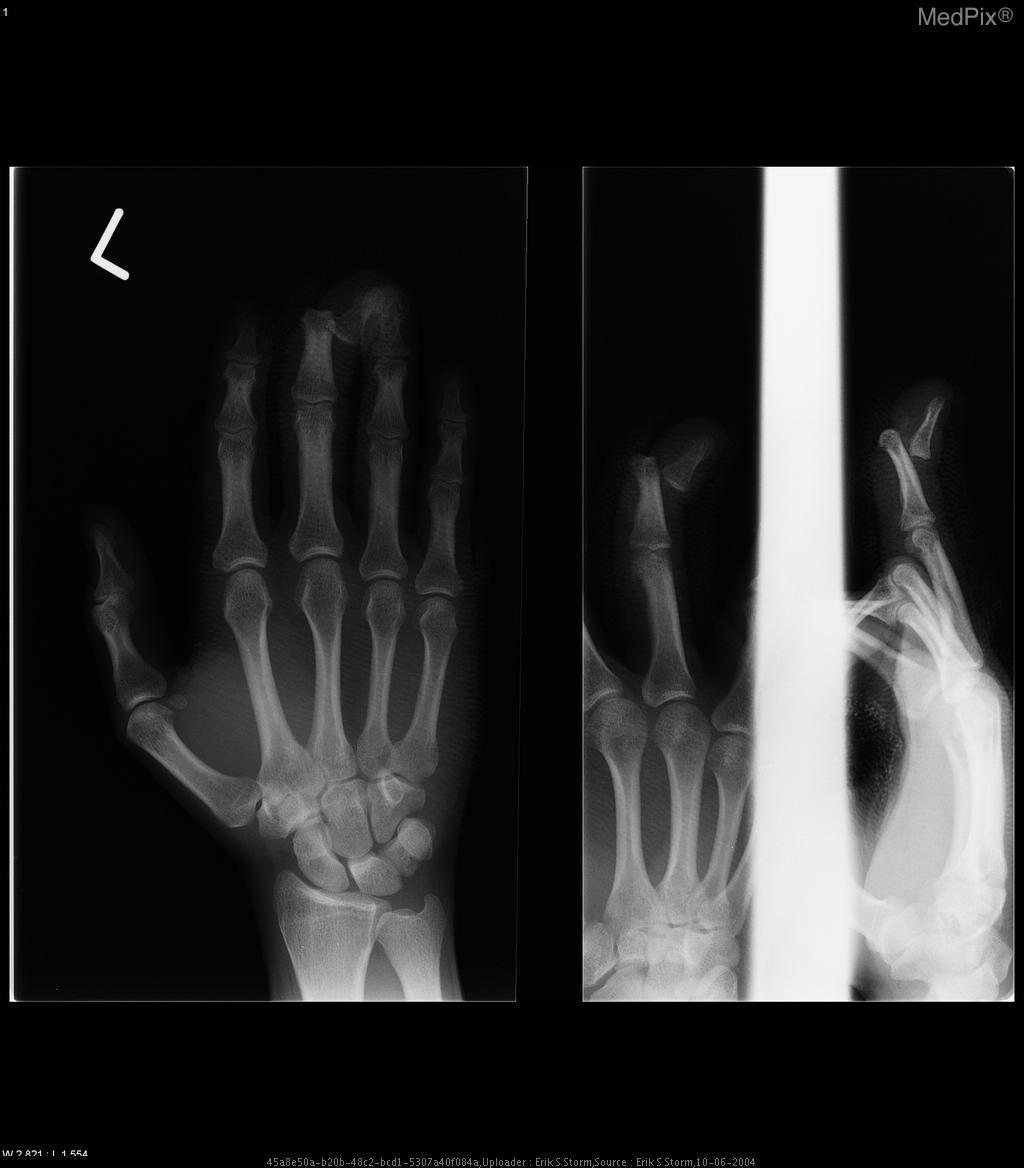}
\end{minipage} &
\begin{minipage}[t]{\linewidth}\vspace{0pt}
\textbf{Question:}
Three views of the left hand clearly demonstrate
the dislocation. Note the protuberance of the
head of the proximal phalynx through the volar
soft tissues of the digit.\\[0.3em]
\textbf{GT:} Traumatic Phalanx Dislocation\\[0.3em]
\textbf{Top-5 predictions:}\\
1.\ Traumatic Phalanx Dislocation (23.5\%)
$\checkmark$\\
2.\ Bizarre Parosteal Osteochondromatous
Proliferation (BPOP) (16.8\%)\\
3.\ Osteopoikilosis (12.6\%)\\
4.\ Dislocation at the interphalangeal joint
of the thumb with avulsion fracture (7.6\%)\\
5.\ Pathologic Fracture in Giant Cell Tumor
(6.6\%)
\end{minipage} \\
\bottomrule
\end{tabular}

\caption{
Qualitative predictions on four radiographs
from the MedPix-derived closed-set VQA benchmark.
Examples A--C share the same caption-derived
question and diagnosis but correspond to
different radiographs. All four examples have
correct Top-1 predictions, with varying
confidence. The percentages indicate the
classifier's predicted probabilities.
}
\label{fig:medpix_vqa_qualitative}
\end{figure*}

\Cref{fig:medpix_vqa_qualitative} presents four radiographs with correct Top-1 predictions but different confidence levels. The first three radiographs share the same caption-derived question and ground-truth diagnosis of chondrocalcinosis, with Top-1 probabilities of 78.3\%, 84.0\%, and 84.7\%, respectively. The fourth example, involving traumatic phalanx dislocation, is also correctly classified but has a lower Top-1 probability of 23.5\%. These examples illustrate variation in prediction confidence within the closed-set task rather than distinct levels of diagnostic reasoning.

This evaluation remains limited by the small dataset and the use of existing case descriptions and diagnostic titles as question--answer pairs. Furthermore, the VQA adapter operates on visual patch tokens without explicit joint or bone localization. The predicted probabilities and cross-attention maps do not establish whether the model relies on anatomically appropriate evidence. Accordingly, these qualitative results illustrate the behavior of the multimodal adapter but do not demonstrate open-ended clinical reasoning or provide causal explanations of its predictions.

\section{Dataset Details} 
\label{supp:dataset_details}

Publicly accessible hand-radiograph datasets remain relatively limited compared with large-scale natural-image and chest-radiograph resources. We curated approximately 30K hand radiographs from six publicly available datasets, encompassing skeletal development, rheumatoid arthritis, musculoskeletal abnormalities, and image-associated clinical descriptions. These resources cover nearly all major public hand-radiograph datasets identified as relevant to the tasks investigated here. Bringing them together broadens the variation in anatomy, patient populations, and imaging conditions available for representation learning. HandXFM leverages this curated collection to learn transferable visual representations without requiring task-specific labels during pretraining.

\Cref{tab:dataset_allocation} distinguishes images used without labels for HandXFM pretraining from the labeled examples used to fit each downstream task and those reserved for evaluation. The unlabeled pretraining set is constructed only from the designated training partitions of RSNA, RHPE, MURA Hand, AIRINIIReumHands, Anhui RA, and MedPix Hand. Reusing a training image during downstream fine-tuning is intentional; \emph{held-out evaluation images are not used in either pretraining or downstream fitting}. The Digital Hand Atlas (DHA) is excluded entirely until external evaluation. 

\begin{table*}[t]
\centering
\caption{Allocation of source datasets to
unlabeled HandXFM pretraining, downstream
model fitting, and held-out evaluation.
Newly annotated localization images follow
the official RSNA partitions.}
\label{tab:dataset_allocation}

\footnotesize
\setlength{\tabcolsep}{4pt}
\renewcommand{\arraystretch}{1.35}

\begin{tabular}{@{}
>{\raggedright\arraybackslash}p{0.16\textwidth}
>{\raggedright\arraybackslash}p{0.17\textwidth}
>{\raggedright\arraybackslash}p{0.19\textwidth}
>{\raggedright\arraybackslash}p{0.19\textwidth}
>{\raggedright\arraybackslash}p{0.22\textwidth}
@{}}
\toprule
\textbf{Dataset}
& \textbf{Downstream task}
& \textbf{Unlabeled pretraining}
& \textbf{Downstream fitting}
& \textbf{Held-out evaluation / separation} \\
\midrule

RSNA Bone Age
& Bone-age regression
& 12,611 training images
& Official training partition
& 1,425 labeled validation images; 200 unlabeled challenge test images not evaluated. Official partitions. \\

\midrule

RHPE
& Bone-age regression
& 5,487 training images
& Training partition
& 713 validation; 79 test images. Disjoint splits. \\

\midrule

MURA Hand
& Abnormality classification
& 5,543 training images
& Training partition
& 460 validation images. Official study-disjoint partition. \\

\midrule

AIRINIIReumHands
& Joint-level SvH regression
& 240 training images
& 240 training images
& 60 validation; 30 test images. Disjoint splits. \\

\midrule

Anhui RA
& Whole-hand SvH regression
& 2,700 training images
& 2,700 training images
& 760 validation; 358 test images. Image-level split. \\

\midrule

MedPix Hand
& Closed-set VQA
& 97 training images
& 97 training image--text pairs
& 11 held-out test pairs. Test images excluded. \\

\midrule

DHA
& External bone-age transfer
& \textbf{Not used}
& \textbf{Not used}; RSNA-trained predictor transferred directly
& 1,103 external images. Entirely excluded. \\

\midrule

Joint detection (ours)
& 17-class localization
& Already included through RSNA training images
& RSNA-train annotations
& RSNA-validation annotations; 6,130 images annotated in total. RSNA partitions. \\

\midrule

Bone segmentation (ours)
& 19-class segmentation
& Already included through RSNA training images
& RSNA-train annotations
& RSNA-validation annotations; 549 images annotated in total. RSNA partitions. \\

\bottomrule
\end{tabular}
\end{table*}

The allocation is defined before downstream training. We preserve the source study- or patient-level grouping when the corresponding identifiers or official grouping information are available, notably for MURA. For collections without reliable patient identifiers, such as the supplied Anhui image split, patient-level duplication cannot be fully ruled out by split membership alone. Checksum-based comparison of image manifests is used to screen for identical files across pretraining and held-out evaluation partitions; this does not establish the absence of all near-duplicates or unidentified repeated patients. The reported dataset sizes and separation protocol are therefore interpreted at the level supported by the available metadata.

\subsection{Public Datasets} \label{supp:public}

\subsubsection{RSNA Bone Age Challenge.}\label{supp:rsna}
The RSNA Pediatric Bone Age dataset consists of left-hand radiographs with skeletal-age labels in months and recorded sex information. The reference age is based on multiple radiologist readings, including repeat assessment, combined through a bias-adjusted weighting procedure. We use its labeled training partition for unlabeled pretraining and for downstream bone-age fitting. The labeled validation partition is reserved for model evaluation; the challenge test images do not provide public reference labels and are not scored in our experiments. The same source partitions also determine the separation of our newly annotated detection and segmentation examples (\Cref{supp:new}).

\begin{table}[t]
\centering
\caption{RSNA bone-age partition sizes and reported mean skeletal age. The challenge test labels are unavailable for the evaluations in this paper.}
\label{tab:rsna_split}

\normalsize
\setlength{\tabcolsep}{3pt}
\renewcommand{\arraystretch}{1.55}

\begin{tabular*}{\linewidth}{@{\extracolsep{\fill}}lccc@{}}
\toprule
\textbf{Split}
& \textbf{\#Images}
& \textbf{Mean Age (mo)}
& \textbf{Labels} \\
\midrule
Training   & 12{,}611 & 127 & Yes \\
Validation & 1{,}425  & 127 & Yes \\
Test       & 200      & 132 & No  \\
\bottomrule
\end{tabular*}
\end{table}

The available demographic summary for the training images is reproduced in \Cref{tab:rsna_demo}. These counts describe radiographs in the provided release; they should not be interpreted as an independently verified count of unique patients.

\begin{table}[t]
\centering
\caption{RSNA bone-age training-image demographics, using the reported sex and age categories.}
\label{tab:rsna_demo}

\normalsize
\setlength{\tabcolsep}{18pt}
\renewcommand{\arraystretch}{1.45}

\begin{tabular}{@{}lc@{}}
\toprule
\textbf{Metric} & \textbf{Value} \\
\midrule
Training images       & 12{,}611 \\
Male-labeled images   & 6{,}833 \\
Female-labeled images & 5{,}778 \\
Mean Age (years)      & 10.61 \\
0--5 years            & 801 \\
6--10 years           & 3{,}487 \\
11--15 years          & 7{,}224 \\
16--20 years          & 1{,}099 \\
\bottomrule
\end{tabular}
\end{table}

\subsubsection{Radiological Hand Pose Estimation (RHPE).} \label{supp:rhpe}
RHPE contains pediatric hand radiographs acquired in Colombia, with sex and bone-age annotations. The release includes both left- and right-hand images and additionally provides anatomical hand-pose landmarks. For HandXFM, the image-level bone-age labels support regression transfer, while only the training radiographs enter the unlabeled pretraining pool. Validation and test images remain excluded from pretraining. The demographic and partition summaries appear in \Cref{tab:rhpe_split,tab:rhpe_demo}.

\begin{table}[t]
\centering
\caption{RHPE partition sizes and reported mean bone ages.}
\label{tab:rhpe_split}

\normalsize
\setlength{\tabcolsep}{3pt}
\renewcommand{\arraystretch}{1.55}

\begin{tabular*}{\linewidth}{@{\extracolsep{\fill}}lccc@{}}
\toprule
\textbf{Split} & \textbf{\#Images} & \textbf{Mean Age (mo)} & \textbf{Labels} \\
\midrule
Training   & 5{,}487 & 126 & Yes \\
Validation & 713     & 126 & Yes \\
Test       & 79      & 126 & Yes \\
\bottomrule
\end{tabular*}
\end{table}

\begin{table}[t]
\centering
\caption{Reported demographics of RHPE training radiographs.}
\label{tab:rhpe_demo}

\normalsize
\setlength{\tabcolsep}{18pt}
\renewcommand{\arraystretch}{1.45}

\begin{tabular}{@{}lc@{}}
\toprule
\textbf{Metric} & \textbf{Value} \\
\midrule
Training images       & 5{,}487 \\
Male-labeled images   & 2{,}524 \\
Female-labeled images & 2{,}963 \\
Mean Age (years)      & 10.5 \\
0--5 years            & 340 \\
6--10 years           & 1{,}590 \\
11--15 years          & 2{,}720 \\
16--20 years          & 837 \\
\bottomrule
\end{tabular}
\end{table}

The RHPE hand-pose annotations were generated through a multi-annotator protocol for 17 landmarks and are not used as supervision in HandXFM pretraining. This distinction matters because the model is pretrained on image pixels without the bone-age or landmark labels, even when the original source dataset makes those annotations available.

\subsubsection{Digital Hand Atlas (DHA).} \label{supp:dha}
DHA provides pediatric left-hand radiographs with skeletal-age assessments and demographic information collected to represent different population groups. We use \textbf{only the first collection cycle of 1,103 images} for external bone-age evaluation. DHA contributes neither unlabeled pretraining images nor downstream training samples: the bone-age model trained on RSNA is applied to DHA without DHA-specific adaptation. The original atlas also includes a subsequent 287-image collection cycle, which is listed for dataset context in \Cref{tab:dha_split} but is not part of the reported external benchmark.

\begin{table}[t]
\centering
\caption{DHA collection cycles. Only the first cycle is used for the reported external-transfer experiment.}
\label{tab:dha_split}

\normalsize
\setlength{\tabcolsep}{3pt}
\renewcommand{\arraystretch}{1.60}

\begin{tabular*}{\linewidth}{@{\extracolsep{\fill}}lccc@{}}
\toprule
\textbf{Cycle}
& \textbf{\#Images}
& \shortstack{\textbf{Age range}\\\textbf{(years)}}
& \shortstack{\textbf{Readings}\\\textbf{per image}} \\
\midrule
First Cycle  & 1{,}103 & 1--18 & 2 \\
Second Cycle & 287     & 5--14 & 4 \\
\bottomrule
\end{tabular*}
\end{table}

\begin{table}[t]
\centering
\caption{Demographic and image-format information for the evaluated DHA first-cycle collection.}
\label{tab:dha_demo}

\normalsize
\setlength{\tabcolsep}{18pt}
\renewcommand{\arraystretch}{1.45}

\begin{tabular}{@{}lc@{}}
\toprule
\textbf{Metric} & \textbf{Value} \\
\midrule
Total Patients       & 1{,}103 \\
Male Patients        & 553 \\
Female Patients      & 550 \\
Ethnic Groups        & CA, AA, HI, AS \\
Mean Age (years)     & 9.5 \\
Age Range            & 1--18 \\
Image Resolution     & 2K $\times$ 2K \\
Format               & DICOM \\
\bottomrule
\end{tabular}
\end{table}

The population and acquisition differences between DHA and RSNA make DHA a useful test of transfer beyond the distributions contributing to training. Its data are strictly external to all HandXFM model fitting.

\subsubsection{AIRINIIReumHands.} \label{supp:airi}
AIRINIIReumHands contains 330 bilateral hand radiographs, including 42 healthy controls and 288 rheumatoid-arthritis cases. The dataset reports joint-level erosion and joint-space narrowing (JSN) scores derived from multiple expert readers, with an average patient age of $48\pm14$ years. The modified Sharp-style annotation scheme considers 17 erosion sites and 19 JSN sites per hand (\Cref{tab:airinii_sites}). HandXFM uses 240 training images without their labels in pretraining; the same partition supplies downstream training labels, with 60 validation and 30 test images kept separate. Its small supervised training size motivates the five-seed stability analysis in the Results Details appendix.

\begin{table}[t]
\centering
\caption{Scored anatomical sites in AIRINIIReumHands (per hand).}
\label{tab:airinii_sites}

\normalsize
\setlength{\tabcolsep}{4pt}
\renewcommand{\arraystretch}{1.50}

\begin{tabular}{@{}p{0.78\linewidth}c@{}}
\toprule
\textbf{Regions} & \textbf{\#Sites} \\
\midrule

Erosion: Interphalangeal, 4~DIP, 4~PIP, 5~MCP,
1st~metacarpal~base, wrist, ulnar, radiocarpal
& 17 \\

\midrule

JSN: Interphalangeal, 4~DIP, 4~PIP, 5~MCP,
CMC~(3--5), wrist, radiocarpal
& 19 \\

\bottomrule
\end{tabular}
\end{table}

The released scores use a Dawid--Skene aggregation approach to combine annotator responses. Joint-region bounding boxes are available for a subset of 268 radiographs; these source annotations are distinct from the RSNA-based 17-class localization dataset created in this work. We retain the source-dataset characteristics in \Cref{tab:airinii_stats} to distinguish its annotation protocol from our own.

\begin{table}[t]
\centering
\caption{Clinical and labeling characteristics reported for AIRINIIReumHands.}
\label{tab:airinii_stats}

\normalsize
\setlength{\tabcolsep}{4pt}
\renewcommand{\arraystretch}{1.40}

\begin{tabular}{@{}
p{0.39\linewidth}
p{0.55\linewidth}
@{}}
\toprule
\textbf{Metric} & \textbf{Value} \\
\midrule

Total Patients
& 330 \\

Healthy Controls
& 42 \\

RA Patients
& 288 \\

Average Age (years)
& $48 \pm 14$ \\

Acquisition Period
& 2019--2022 \\

Hand Views
& Bilateral (1 radiograph per subject) \\

Annotation Experts
& 3 radiologists (5, 7, and 30~years of experience) \\

Scoring Method
& Modified Sharp (erosion: 0--5, JSN: 0--4) \\

Consensus Algorithm
& Dawid--Skene aggregation \\

Joint Locations Annotated
& 21 per hand (bounding boxes) \\

\bottomrule
\end{tabular}
\end{table}

\subsubsection{Anhui RA Dataset.} \label{supp:anhui}
The Anhui RA collection contains 3,818 bilateral-hand examinations annotated for rheumatoid-arthritis damage severity. Radiologist-derived erosion, JSN, and overall Sharp--van der Heijde (SvH) scores provide supervision for the whole-hand regression task. Only the \textbf{2,700 training images} are included in HandXFM pretraining. We use the corresponding image-level split of 760 validation and 358 test images for model assessment; the three partitions sum to 3,818. The provided release does not support an independent audit of patient-level overlap, which is a limitation of this split.

\begin{table}[t]
\centering
\caption{Anhui RA image-level partitions used for pretraining and SvH regression.}
\label{tab:anhui_split}

\normalsize
\setlength{\tabcolsep}{3pt}
\renewcommand{\arraystretch}{1.55}

\begin{tabular*}{\linewidth}{@{\extracolsep{\fill}}lclc@{}}
\toprule
\textbf{Split}
& \textbf{\#Images}
& \textbf{Annotations}
& \textbf{Method} \\
\midrule

Training
& 2{,}700
& \shortstack[l]{JSN, Erosion,\\Sharp score}
& SvH \\

Validation
& 760
& \shortstack[l]{JSN, Erosion,\\Sharp score}
& SvH \\

Test
& 358
& \shortstack[l]{JSN, Erosion,\\Sharp score}
& SvH \\

\bottomrule
\end{tabular*}
\end{table}

\begin{table}[t]
\centering
\caption{Anhui RA source-dataset characteristics and the train/validation/test partition counts.}
\label{tab:anhui_stats}

\normalsize
\setlength{\tabcolsep}{3pt}
\renewcommand{\arraystretch}{1.40}

\begin{tabular}{@{}
p{0.40\linewidth}
p{0.55\linewidth}
@{}}
\toprule
\textbf{Metric} & \textbf{Value} \\
\midrule

Total Radiographs
& 3{,}818 \\

Institution
& Anhui University of Traditional Chinese Medicine \\

Acquisition Period
& 2020--2022 \\

Average Age (years)
& Not reported \\

Annotation Experts
& 2 radiologists \\

Scoring System
& Sharp--van der Heijde (SvH) \\

Labeled Attributes
& JSN, Erosion, Sharp Score \\

Split Ratio (Train/Val/Test)
& 2700 / 760 / 358 \\

Image Format
& X-ray (DICOM) \\

Diagnosis Types
& RA or suspected RA \\

\bottomrule
\end{tabular}
\end{table}

\subsubsection{MURA Hand Subset.}
\label{supp:mura}

We select the \texttt{XR\_HAND} category from the
MURA musculoskeletal radiography release for
normal-versus-abnormal classification. The
selected subset contains 6,003 radiographs,
including 5,543 training images and 460
validation images (\Cref{tab:mura_split}).
A study may contain multiple radiographic
views. HandXFM pretraining uses only images
from the designated training partition,
while the held-out validation partition is
reserved for abnormality classification.


\begin{table}[t]
\centering
\caption{MURA hand-subset image statistics
for the training and validation partitions.}
\label{tab:mura_split}

\normalsize
\setlength{\tabcolsep}{3pt}
\renewcommand{\arraystretch}{1.55}

\begin{tabular*}{\linewidth}{@{\extracolsep{\fill}}lcccc@{}}
\toprule
\textbf{Split}
& \textbf{Images}
& \textbf{Normal}
& \textbf{Abnormal}
& \shortstack{\textbf{Normal}\\\textbf{ratio}} \\
\midrule

Training   & 5{,}543 & 4{,}059 & 1{,}484 & 0.732 \\
Validation & 460     & 271     & 189     & 0.589 \\

\midrule

\textbf{Total}
& \textbf{6{,}003}
& \textbf{4{,}330}
& \textbf{1{,}673}
& \textbf{0.721} \\

\bottomrule
\end{tabular*}
\end{table}

\subsubsection{MedPix Dataset.} \label{supp:medpix}
The MedPix teaching-case archive provides radiographic images linked to diagnostic titles and descriptive text~\cite{medpix}. We select 108 hand X-ray image--text examples and create a closed-set evaluation by treating the descriptive caption as a question-like textual input and the diagnostic title as the answer label. The resulting split contains \textbf{97 training pairs and 11 held-out test pairs}. Only training images participate in HandXFM's unlabeled pretraining, and no test image is reused in model fitting. Since the questions are derived from existing case descriptions rather than independently authored clinical queries, this benchmark measures constrained multimodal transfer and should not be regarded as open-ended clinical VQA.

\subsection{Newly Annotated Datasets} \label{supp:new}

\paragraph{Joint Detection Dataset.} \label{supp:joint_det_data}
We construct a 17-category joint/landmark detection benchmark by annotating bounding boxes in 6,130 RSNA radiographs. The classes cover DIP joints of digits 2--5, proximal interphalangeal joints of digits 2--5 and the thumb interphalangeal joint (\texttt{PIP\_1} in the dataset label scheme), MCP joints of digits 1--5, two distal forearm landmarks, and the wrist center. 
The 17 target categories comprise the DIP,
PIP/IP, and MCP joints, together with distal
radius, distal ulna, and wrist-center landmarks.
The corresponding class names and anatomical
definitions are listed in
\Cref{tab:joint_definitions}.

Annotation images were sourced from the official RSNA train and validation partitions, at approximately a 9:1 ratio. Detector fitting uses only labels from the RSNA training partition, and evaluation uses annotations on RSNA validation images. Since unlabeled pretraining likewise excludes RSNA validation images, no detection-evaluation image is included in HandXFM pretraining or detector training.

\begin{table}[t]
\centering
\caption{Model-label mapping for the 17 RSNA-derived joint/landmark detection categories.}
\label{tab:joint_definitions}

\normalsize
\setlength{\tabcolsep}{3pt}
\renewcommand{\arraystretch}{1.30}

\begin{tabular}{@{}
p{0.20\linewidth}
p{0.76\linewidth}
@{}}
\toprule
\textbf{Joint label} & \textbf{Anatomical definition} \\
\midrule
DIP\_5 & Distal interphalangeal joint of the little finger \\
DIP\_4 & Distal interphalangeal joint of the ring finger \\
DIP\_3 & Distal interphalangeal joint of the middle finger \\
DIP\_2 & Distal interphalangeal joint of the index finger \\
PIP\_5 & Proximal interphalangeal joint of the little finger \\
PIP\_4 & Proximal interphalangeal joint of the ring finger \\
PIP\_3 & Proximal interphalangeal joint of the middle finger \\
PIP\_2 & Proximal interphalangeal joint of the index finger \\
PIP\_1 & Interphalangeal joint of the thumb (dataset label \texttt{PIP\_1}) \\
MCP\_5 & Metacarpophalangeal joint of the little finger \\
MCP\_4 & Metacarpophalangeal joint of the ring finger \\
MCP\_3 & Metacarpophalangeal joint of the middle finger \\
MCP\_2 & Metacarpophalangeal joint of the index finger \\
MCP\_1 & Metacarpophalangeal joint of the thumb \\
Radius & Distal tip of the radius bone \\
Ulna & Distal tip of the ulna bone \\
Wrist & Center point of the carpal (wrist) region \\
\bottomrule
\end{tabular}
\end{table}

\paragraph{Bone Segmentation Dataset.} \label{supp:bone_seg_data}

We annotate 549 RSNA radiographs with
pixel-level masks for 19 individual
metacarpal and phalangeal categories.
Annotations from the RSNA training partition
are used for downstream model fitting,
whereas those from the validation partition
are reserved for evaluation. The complete
class definitions and output-label mapping
are provided in \Cref{tab:bone_definitions}.

\begin{table}[t]
\centering
\caption{Output indices for the 19 anatomically distinct bone-segmentation classes.}
\label{tab:bone_definitions}

\normalsize
\setlength{\tabcolsep}{3pt}
\renewcommand{\arraystretch}{1.25}

\begin{tabular}{@{}
p{0.20\linewidth}
p{0.76\linewidth}
@{}}
\toprule
\textbf{Label ID} & \textbf{Anatomical definition} \\
\midrule
0  & Metacarpal bone of the index finger \\
1  & Metacarpal bone of the middle finger \\
2  & Metacarpal bone of the thumb \\
3  & Metacarpal bone of the ring finger \\
4  & Metacarpal bone of the little finger \\
5  & Proximal phalanx of the middle finger \\
6  & Proximal phalanx of the index finger \\
7  & Proximal phalanx of the ring finger \\
8  & Proximal phalanx of the thumb \\
9  & Proximal phalanx of the little finger \\
10 & Middle phalanx of the middle finger \\
11 & Middle phalanx of the ring finger \\
12 & Middle phalanx of the index finger \\
13 & Distal phalanx of the thumb \\
14 & Middle phalanx of the little finger \\
15 & Distal phalanx of the middle finger \\
16 & Distal phalanx of the ring finger \\
17 & Distal phalanx of the index finger \\
18 & Distal phalanx of the little finger \\
\bottomrule
\end{tabular}
\end{table}

\paragraph{Annotation Standards.} \label{supp:anno}
Our anatomical class definitions follow \emph{Terminologia Anatomica} (2nd edition) and are cross-checked against RSNA RadLex terminology. Digits are indexed 1 through 5 from thumb to little finger. The anatomical naming convention requires a small distinction from the stored detection class name: \texttt{PIP\_1} refers to the thumb's single interphalangeal joint, not to a separate proximal interphalangeal joint. The distal-radius, distal-ulna, and wrist-center classes are localization landmarks rather than additional finger joints. For segmentation, each integer class ID from 0 to 18 corresponds to one specific bone, as defined in \Cref{tab:bone_definitions}.

Two annotators with musculoskeletal training participated in constructing and reviewing the new annotations. Initial bounding boxes and masks were produced by one annotator using Roboflow, with subsequent joint review and consensus resolution. An automated quality-assurance screen flagged incomplete cases, missing or repeated class labels, unexpected category counts, questionable hand orientation, and anatomically implausible spatial arrangements; flagged annotations were reviewed before finalization. A further audit comprised 1,000 randomly sampled detection images and 100 segmentation masks, reviewed by both annotators for landmark placement, mask alignment, and correct category assignment. Because the annotation process used consensus review rather than two independent full annotation sets, inter-annotator IoU or Cohen's $\kappa$ cannot be reported as independent agreement measures.

\section{Limitations and Future Work}
\label{supp:limitations_future}

Despite its performance across heterogeneous downstream tasks, HandXFM has several limitations related to expert supervision, data coverage, and downstream adaptation.

\paragraph{Expert fusion formulation.}
HandXFM combines the latent targets of two frozen visual experts through a shared alignment and reconstruction framework. Although the teacher-composition ablations demonstrate complementary transfer across tasks, the current formulation assigns equal importance to the two experts rather than adapting their contributions to individual images or anatomical regions. Moreover, the relative loss weights were selected based on preliminary loss-scale observations rather than a comprehensive hyperparameter sensitivity study. The objective ablations establish the contribution of alignment and reconstruction, but do not characterize sensitivity to all possible weighting configurations. Future work could investigate adaptive expert weighting and alternative fusion mechanisms, particularly for tasks with different requirements for global semantic information and localized anatomical detail.

\paragraph{Scope of pretraining data.}
We curated approximately 30K hand radiographs from six publicly available datasets, covering multiple clinical tasks and imaging conditions. Nevertheless, the available data remain limited in comparison with larger medical imaging domains. In particular, uncommon abnormalities, advanced disease manifestations, and variations in acquisition protocols may not be sufficiently represented. The diversity of the training sources also does not guarantee equivalent performance across institutions or patient subgroups. Additional datasets encompassing broader patient populations, disease severity, and imaging conditions would enable more extensive assessment of representation transfer and robustness.

\paragraph{Downstream adapter design.}
HandXFM is evaluated using relatively lightweight task-specific heads to assess the transferability of its pretrained representations. However, these adapters may not fully exploit the anatomical information encoded by the backbone. For example, the VQA adapter applies cross-attention over visual patch tokens without explicitly incorporating detected joints or segmented bones. This can limit its ability to distinguish abnormalities requiring precise assessment of local geometry, cortical continuity, or relationships between anatomical structures.

The MedPix-derived VQA evaluation is also limited by its small sample size and closed-set answer formulation. Consequently, performance on this benchmark does not establish open-ended clinical reasoning capability. Similarly, the cross-attention visualizations provide qualitative examples of text--image correspondence rather than validated causal explanations of model predictions. Future work could investigate anatomy-conditioned multimodal adapters and larger benchmarks with region-level text annotations to evaluate visual grounding more rigorously.

\section{Reproducibility Check}
\label{supp:reproducibility}

We summarize the computational environment, data processing procedures, and experimental configurations used for HandXFM pretraining and downstream evaluation. Additional architecture and optimization details are provided in \Cref{supp:pretrain_config}, while dataset-specific partitions are described in \Cref{supp:dataset_details}.

\subsection{Compute}
\label{supp:repro_compute}

\paragraph{Hardware.}
Experiments were conducted on a Linux server equipped with an NVIDIA A100 GPU and an Intel Xeon Platinum 8480+ CPU.

\paragraph{Software.}
HandXFM is implemented in PyTorch. The implementation uses Hugging Face Transformers for the student backbone and radiography-domain teacher, \texttt{open\_clip} for BiomedCLIP, and \texttt{timm} and \texttt{torchvision} for model components and image processing.

\paragraph{Reproducibility controls.}
For each downstream benchmark, all compared models use the designated training and evaluation partitions and the same task-specific metric definitions. Model initialization, data processing, optimization, and masking configurations are specified independently of the evaluation data. The multi-seed experiments on AIRINIIReumHands additionally quantify variability across different random initializations, as detailed in \Cref{supp:appendix_multiseed_airinii}.

\subsection{Data organization and preprocessing}
\label{supp:repro_data}

\paragraph{Data roots.}
Pretraining images are organized by dataset source and loaded from configurable root directories. The image loader supports commonly used radiographic image formats, including PNG, JPEG, and TIFF. Dataset-specific training partitions determine which images are eligible for pretraining. The allocation of images across pretraining, downstream training, and evaluation is summarized in \Cref{tab:dataset_allocation}.

\paragraph{Transforms.}
Pretraining applies resizing, random horizontal flipping, rotation, and center cropping to obtain $224\times224$ inputs. Pixel intensities are converted to $[0,1]$, followed by the preprocessing required by each pretrained encoder. To avoid repeated input scaling, the DINOv3 image processor uses \texttt{do\_rescale=False}. The complete augmentation settings are reported in \Cref{supp:augmentation}.

\paragraph{Pretraining split policy.}
HandXFM uses only the designated training partitions of the six source datasets for unlabeled pretraining. Validation and test images are excluded, and the Digital Hand Atlas is held out for external bone-age evaluation. For the RSNA-derived joint-detection and bone-segmentation benchmarks, downstream training annotations originate from the RSNA training partition, whereas evaluation annotations are drawn from the disjoint validation partition. The corresponding split definitions and overlap controls are documented in \Cref{tab:dataset_allocation}.

\subsection{Training configuration}
\label{supp:repro_train}

\paragraph{Optimization.}
Pretraining uses AdamW with separate learning rates for the student encoder and auxiliary modules, including the latent regressor and expert projection layers. The student is frozen during the initial two epochs and subsequently optimized jointly with the auxiliary modules. The optimizer settings, learning rates, batch size, and training schedule are specified in \Cref{tab:handxfm_pretrain_config}. Checkpoints are saved periodically throughout pretraining.

\paragraph{Masking and losses.}
The training configuration specifies the active expert encoders, masking strategy, masking ratio, and coefficients of the alignment, reconstruction, CLS, and EMA consistency terms. The main configuration employs edge-aware masking with a ratio of $0.5$ and an edge-mixing coefficient of $0.5$. Ablation variants modify the corresponding components while retaining the remaining training settings, as described in \Cref{supp:variant}.

\subsection{Code entry points and artifacts}
\label{supp:repro_code}

\paragraph{Code entry points.}
The pretraining pipeline is configured through runtime arguments specifying dataset paths, pretrained checkpoints, expert selection, masking settings, loss coefficients, optimizer parameters, and output directories. Downstream experiments use task-specific training and evaluation procedures with the corresponding pretrained encoder weights. The configurations for the main model and pretraining variants follow the settings reported in \Cref{supp:pretrain_config}.

\paragraph{Saved artifacts.}
Training checkpoints contain the student and EMA encoder states, masked latent regressor, expert projection modules, optimizer state, and training progress information. The EMA encoder from the final pretraining checkpoint is used as the default HandXFM backbone for downstream adaptation. Training logs record the optimization process, including the individual loss components, while downstream evaluation produces the task-specific metrics reported in the main paper and supplementary experiments.

\paragraph{Code and checkpoint availability.}
The HandXFM pretraining implementation,
dependencies, and setup instructions are
publicly available at
\url{https://github.com/YUESAI/HandXFM}.
The pretrained visual encoder, model
configuration, and usage documentation are
available at
\url{https://huggingface.co/Yuesai/HandXFM}.

\section{Broader Impact}
\label{supp:broader}

HandXFM provides a reusable visual representation for hand radiographs, with potential applications in skeletal maturity estimation, rheumatoid arthritis assessment, abnormality screening, and anatomical localization. By transferring knowledge from existing biomedical and radiography-domain models, the framework may reduce the need to develop separate visual representations for individual hand X-ray tasks. This could facilitate research on musculoskeletal imaging applications where task-specific annotated data are limited.

However, HandXFM is a research model and has not been established as a standalone clinical diagnostic system. Its performance may be affected by differences in patient demographics, disease prevalence, imaging equipment, and acquisition protocols. In particular, underrepresented anatomical abnormalities or patient populations may exhibit different error characteristics from those observed in the evaluated benchmarks. The present results therefore should not be interpreted as evidence of clinical readiness or uniformly reliable performance across patient subgroups.

Responsible clinical translation would require independent external validation, systematic evaluation of demographic and acquisition-related performance differences, and prospective assessment under clinical workflows. Automated predictions should be interpreted with appropriate clinical oversight, particularly when errors could affect diagnostic or treatment decisions.

\clearpage


\newpage

\end{document}